\documentclass[manuscript,screen]{acmart}

\usepackage[most]{tcolorbox}

\usepackage{graphicx}
\usepackage{multirow} 
\usepackage{array}

\usepackage{subfig}  
\usepackage{overpic} 

\AtBeginDocument{%
  \providecommand\BibTeX{{%
    \normalfont B\kern-0.5em{\scshape i\kern-0.25em b}\kern-0.8em\TeX}}}

\setcopyright{acmlicensed}
\copyrightyear{2018}
\acmYear{2018}
\acmDOI{XXXXXXX.XXXXXXX}

\begin{document}

\title{Dynamic NeRF: A Review}


\author{Jinwei Lin}
\authornote{Super Researcher}
\email{Jinwei.Lin@monash.edu}
\orcid{0000-0003-0558-6699}
\affiliation{%
  \institution{Monash University}
  \city{Clayton}
  \state{Victoria}
  \country{Australia}
}


\renewcommand{\shortauthors}{Jinwei Lin, et al.}

\begin{abstract}
Neural Radiance Field(NeRF) is an novel implicit method to achieve the 3D reconstruction and representation with a high resolution. After the first research of NeRF is proposed, NeRF has gained a robust developing power and is booming in the 3D modeling, representation and reconstruction areas. However the first and most of the followed research projects based on NeRF is static, which are weak in the practical applications. Therefore, more researcher are interested and focused on the study of dynamic NeRF that is more feasible and useful in practical applications or situations. Compared with the static NeRF, implementing the Dynamic NeRF is more difficult and complex. But Dynamic is more potential in the future even is the basic of Editable NeRF. In this review, we made a detailed and abundant statement for the development and important implementation principles of Dynamci NeRF. The analysis of main principle and development of Dynamic NeRF is from 2021 to 2023, including the most of the Dynamic NeRF projects. What is more, with colorful and novel special designed figures and table, We also made a detailed comparison and analysis of different features of various of Dynamic. Besides, we analyzed and discussed the key methods to implement a Dynamic NeRF. The volume of the reference papers is large. The statements and comparisons are multidimensional. With a reading of this review, the whole development history and most of the main design method or principles of Dynamic NeRF can be easy understood and gained.  

\end{abstract}




\maketitle

\section{Introduction}

3D reconstruction or the 3D novel views synthesis is the basic research foundation of the Neural Radiance Field (NeRF) Rendering \cite{mildenhall2021nerf}. 3D reconstruction \cite{zollhofer2018state} or 3D novel views synthesis \cite{park2017transformation} are all two significant approaches and research fields in the area of 3D design and 3D modeling of computer vision \cite{fathi2015automated}. 3D reconstruction has been an important developing technology in the recent years \cite{kang2020review}, but the prosperous developing stage of the traditional 3D reconstruction, more researches that are relevant to the traditional 3D reconstruction are gradually tending to application research \cite{ma2018review}. Most of the traditional 3D reconstruction methods are explicit \cite{poursaeed2020coupling}. Due to some historic disadvantages of the traditional 3D reconstruction \cite{fu2021single}, like being weak in the high resolution reconstruction of 3D scenes, needing a large amount of input data and deeply relying on the hardware etc. After \citet{mildenhall2021nerf} proposed the NeRF, which represents the Neural Radiance Field Rendering, a novel research field of implicit 3D reconstruction has been established. The most obvious advantages of NeRF is that the NeRF can make a 3D views synthesis with an extremely high resolution and easily use few image or single-view images to make a complex 3D thesis.

\subsection{Main Content of this Review}
In this review, a comprehensive review of the principles and techniques of Dynamic NeRF will be presented. This review will analyze the whole development history and procedure from the first research result of Dynamic NeRF to the current research of the Dynamic NeRF. In this review, the cause why the Dynamic NeRF will be an inevitable technology will be presented.  The statement route of this review will focus on the principle statement and analysis, statistical analysis and comparison. The review will make a comprehensive introduction and statement for relative background knowledge about the NeRF, and make the detailed technology analysis and discussion of the Dynamic NeRF. The aim of this review is to propose a comprehensive review of the researchers to understand the research development and current novel research direction of the research field of Dynamic NeRF. In the statement, We will analyze the development history and key factors that influence the development of DyNeRF. After the basic analysis and statement for the existing Dynamic Principle and Technologies, we will discuss the potential research directions of Dynamic NeRF in the future.


\subsection{Main Content of this Review}
To propose a novel and valuable research review of the Dynamic NeRF, the statement and analysis will focus on two review dimensions: the horizontal analysis and comparison based on the developing time, and the vertical analysis and comparison based on the same range. Using the review method Horizontal-vertical Analysis Method is the first prominent feature of this review. The second prominent feature of this review is that the number of the reference papers used is abundant, which means we will analyze most of the published papers that are relevant to Dynamic NeRF. The third prominent feature of this review is that there are ample original images and tables throughout the main content, which will give the reader a more intuitive and comprehensive analysis and contrast presentation to understand the research review of Dynamic NeRF better. The fourth prominent feature of this review is that this review will not only analyze the current development principles and technologies of Dynamic NeRF but also analyze and discuss the future potential development direction of Dynamic NeRF according to the sufficient relative meteritals. The fifth prominent feature of this review is that this review will focus the analysis and comparison of the performances achieved by different design methods, principles and algorithms, moreover, this review will chase best performance from different direction in the research field of Dynamic NeRF, which will assist the reader to chase the State-of-the-Art (SOTA) research results of Dynamic NeRF better. Briefly, we will make a detailed enough review that has the abundant data analysis and characteristics comparison from different dimensions to propose a useful research review and research reference for the research of Dynamic NeRF.

\subsection{Research Significance of Dynamic NeRF}
After \citet{mildenhall2021nerf} has proposed the pioneering research of NeRF in 2021, due to the huge potential research value brought by the NeRF, there are various research works that are relevant to original NeRF \cite{mildenhall2021nerf}, such as: Mip-nerf \cite{barron2021mip}, which focuses on designing a NeRF that is available for multiscale representation and anti-aliasing; Block-nerf \cite{tancik2022block}, which focuses on the synthesis of scalable large scene; Mip-nerf 360 \cite{barron2022mip}, which focuses on rendering the panoramic and stereo scenes or objects. However, all of these works are not the Dynamic NeRF work, which means the rendered results of all of these research works are static. The NeRF that renders the static 3D scene or objects is named as Static NeRF.

Compared with the static images, the dynamic video will draw more attention from the audiences. Meanwhile, the dynamic video can contain more information and can be used in more application fields, compared with the static images. In the area of 3D reconstruction or NeRF, analogously, the Dynamic NeRF will represent more abundant information and be used in more application fields, rather than the Static NeRF. 

Compared with the Static NeRF, the dynamic NeRF can represent more information and has a broader application field. Therefore, the future trend of the research of NeRF will focus more on the Dynamic NeRF, which means the research of Dynamic NeRF, in the future, will become more and more important and valuable in the area of NeRF.

\section{Background}
To grasp or comprehend a principle or theory, an efficient method is to understand the background of it. In this section, we will make a detailed statement and analysis of the development history and technology background of Dynamic NeRF. We will use abundant charts and tables that are created following the corresponding statistics and analysis data and design. All charts and tables in this review are original to maintain the novelty of this review paper, and on the other hand, for providing the design thought for the writing of the relative papers, especially the review papers.

\subsection{Development History of Dynamic NeRF}

In this section, we will analyze the  whole development history from the first original research work on the topic of Dynamic NeRF to the current research of the Dynamic NeRF. The narrative structure will follow the HVAM methodology.

\subsubsection{Development History Summary}

\begin{figure}[t]
    \centering
    \includegraphics[width=0.99\columnwidth]{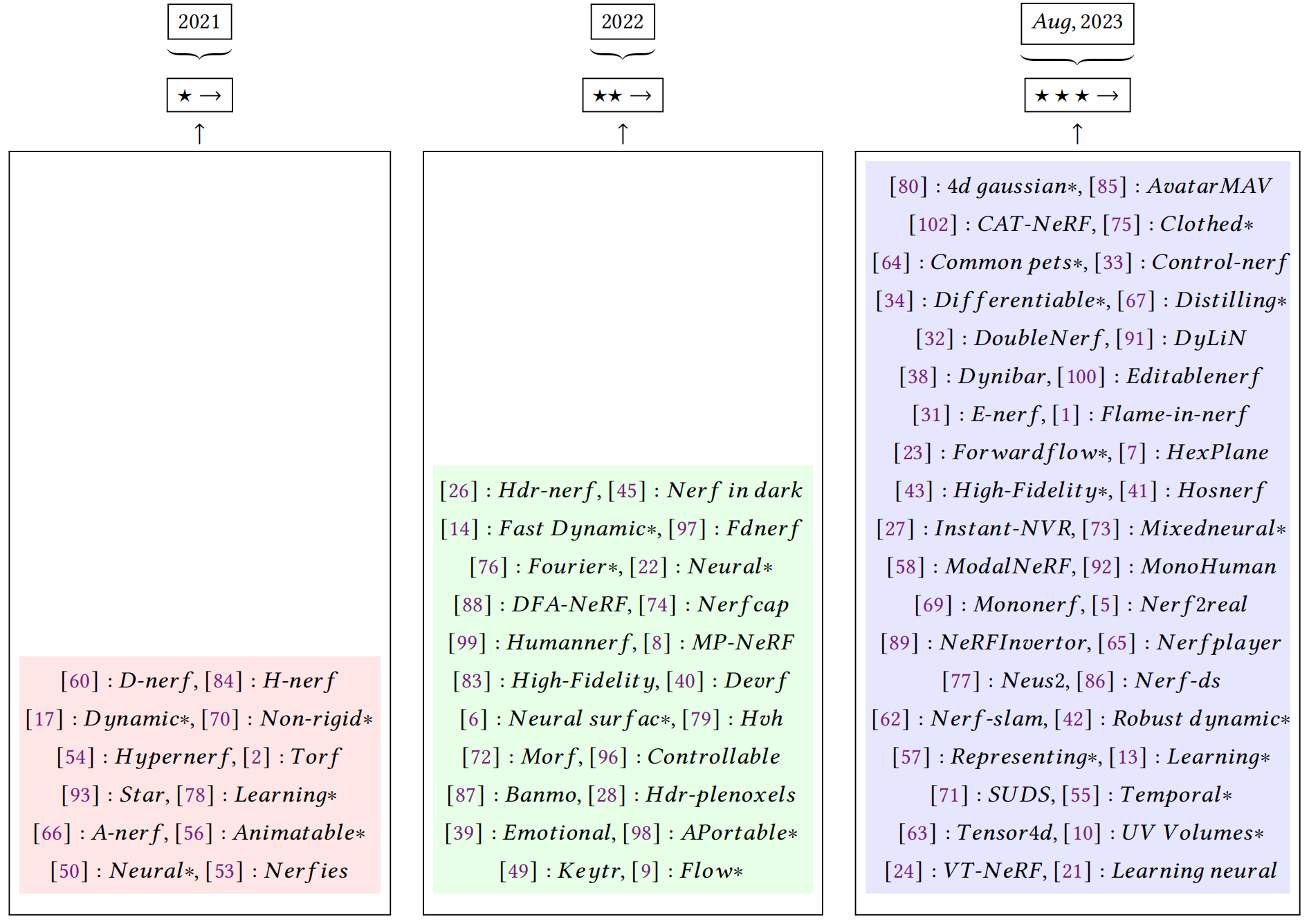}
    \caption{Time Development Chart of Dynamic NeRF.}
    \label{fig0}
\end{figure}

First, we make a global summary of the development of Dynamic NeRF from 2021 to 2023. We design a special chart table to describe them, as shown in Figure \ref{fig0} that is called the Time Development Chart (TDC). TDC is a novel method to describe the development thread of one thing with interactive citation labels. 

As shown in Table \ref{tb1}, the TDC of Dynamic NeRF consists of three main components: the developments of Dynamic NeRF in 2021, 2022 and 2023. For the development component of each year, different color boxes are used to contain the items of research work of the year. In the Figure \ref{fig0}, colors green, yellow and pink are used to represent the content boxes of the years 2021, 2022 and 2023. In the item box, each item is expressed by an item number and the abbreviation of the research work. The abbreviation that is end without the symbol * is the original name of the research work. The abbreviation that is end with the symbol * is the abbreviation of first or second words of the title of the research work. The symbol * represents the rest of the title of the research work.

As shown in Figure \ref{fig0}, in the comparison of the numbers of the research works from 2021 to 2023, the number of the research works of 2021 is smallest, which is the first development year of Dynamic NeRF. In 2022, the number of the research works of Dynamic NeRF was more than two times that of 2021, which means the research of Dynamic NeRF has got a great development in the second year. In 2023, not the whole year yet, to Aug 2023, the number of research of Dynamic NeRF has grown more than the number in 2022, which means the developing speed of the research of the field of Dynamic NeRF has got faster developing trend, which means more and more research attentions have been focused on this field. The reason that the development of the research of Dynamic NeRF becomes more and more important and attention-grabbing, as our consideration, is the target of NeRF technology and principle is to develop a technology that can replace or beyond the performance of the normal explicit rendering technologies and principles, therefore, a dynamic NeRF technology and its implementation theory is an inevitable target achievement and necessary process.

\subsubsection{Development of Dynamic in 2021}

After \citet{mildenhall2021nerf} proposed the first research paper of After \citet{mildenhall2021nerf} proposed the first research paper of NeRF, the research file of NeRF is still focusing on researching the original NeRF that is considered as the static NeRF, until other researchers started to notice the dynamic NeRF, such as \citet{pumarola2021d} propose the D-nerf. D-nerf\cite{mildenhall2021nerf}  has extended the application field of NeRF from a static domain to a dynamic domain, which renders the rigid and non-rigid objects by adding the time as an input parameter. The participation of the parameter time in the rendering function has played an important role, in \cite{xu2021h}, the time is also added as an input parameter, which has also updated the loss function and structured the relative implicit human body model. The rendering result of using the  monocular camera to make NeRF is also an important research field in normal NeRF or Dynamic NeRF. The monocular camera is usually discussed with the binocular camera, which represent one or two cameras respectively. \citet{gao2021dynamic} has researched the performance of Dynamic NeRF with a monocular camera, which can render the dynamic scene by using the differentiable and continuous function, and render the a time-varying dynamic NeRF and the time-invariant static NeRF. For the rendered scene of Dynamic NeRF, can be classified as rigid scene and non-rigid scene, which consist of rigid objects and non-rigid objects respectively. With compassion, the non-rigid objects have more possibilities in more situations to transform when under the external force. \citet{tretschk2021non} has proposed a Dynamic NeRF for non-rigid NeRF, which uses the RGB images as the input data, and uses the ray bending to make the scene transformation to make NeRF dynamic. 

In this stage, the main method to build a Dynamic NeRF is to make the deep learning of the deformation field mapping of the input images coordinates date, then change the deformation field mapping to a coordinate space that has a canonical template. But this kind of change is not the topological change, because the change is continuous and the topological change is discontinuous. To address this problem, \citet{park2021hypernerf} proposed the Hypernerf that can make the Dynamic NeRF in a topological discontinuous field. This kind of method has improved the performance of other dynamic NeRFs. In this research stage, the research applications of using the Dynamic NeRF to handle the view synthesis from the monocular images is available, but there are still difficulties for the Dynamic NeRF being used for big scale scenes. For this situation, \citet{attal2021torf} proposed the Torf to make the synthesis rendering of Dynamic NeRF by replacing the priors with the data from a time-of-flight (ToF) camera. Using the technology of NeRF can predict the information of the objects in the synthesized scene. But this kind of prediction ability will be invalid once the objects in the scene move. \citet{yuan2021star} proposed the STaR, a method of dynamic NeRF that renders the moving rigid object and decomposes the object into two components in the meantime without human supervision.

In the research field of Dynamic NeRF, one important and high-frequency used research object is human, especially the actions and the facial expression of humans. The head expression of the head with emotion usually is used as the object of Dynamic NeRF \cite{wang2021learning}. Not only the head, the whole components of human body shape are important \cite{su2021nerf}, \citet{su2021nerf} proposed the A-nerf to render the dynamic scene of the whole human body shape. Therefore, animatable human body rendering is an important topic in Dynamic NeRF,  \citet{peng2021animatable} also proposed a reconstruction method of dynamic human body model. To make the Dynamic NeRF more efficient, \citet{ost2021neural} proposed a method that can learn the representation of the scene graph to represent the scene that has multiple objects to a scene graph, which can decompose the multiple objects of the scene into individual single objects. As the development of NeRF, the application in mobile and portable devices has drawn more and more attention. \citet{park2021nerfies} proposed Nerfies, the first research work that uses the data from the mobile device to make the Dynamic NeRF by using the optimization that is based on the physical simulating and geometric processing  of an additional volume deformation field that is continuous. 

In the whole stage of 2021, the research of DYnamic NeRF is on the inital stage. Dynamic NeRF can be focused in the same year of the year when the NeRF is proposed first, which means the Dynamic NeRF has drawn more and more attention of the research field of NeRF.

\subsubsection{Development of Dynamic in 2022}

Now comes the development of Dynamic NeRF in 2022, in which stage the development of the Dynamic NeRF has got a better growth, compared with 2021. In this stage, researchers began to put more attention on improving the performance of Dynamic NeRF. \citet{huang2022hdr} proposed the Hdr-nerf that using the modeling of the physical imaging processing of Dynamic NeRF to reconstruct the novel High Dynamic Range (HDR) views and Low Dynamic Range (LDR) views Dynamic NeRF with various exposure parameters. This means the impacts from the different environments of Dynamic NeRF have started to be emphasized. \citet{mildenhall2022nerf} proposed the Nerf in the dark, which focuses on dealing with the Dynamic NeRF in a dark or light-lack environment. 

Now comes the development of Dynamic NeRF in 2022, in which stage the development of the Dynamic NeRF has got a better growth, compared with 2021. In this stage, researchers began to put more attention on improving the performance of Dynamic NeRF. \citet{huang2022hdr} proposed the Hdr-nerf that using the modeling of the physical imaging processing of Dynamic NeRF to reconstruct the novel High Dynamic Range (HDR) views and Low Dynamic Range (LDR) views Dynamic NeRF with various exposure parameters. This means the impacts from the different environments of Dynamic NeRF have started to be emphasized. \citet{mildenhall2022nerf} proposed the Nerf in the dark, which focuses on dealing with the Dynamic NeRF in a dark or light-lack environment. 

Focusing on improving the performance or processing speed of Dynamic NeRF is still an important research direction in this stage. \citet{fang2022fast} proposed a method to improve the processing speed of Dynamic NeRF and synchronously save the calculation resource by using the time-aware voxel feature and coordinate deformation neural network to represent the synthesized scene. To get higher performance of Dynamic NeRF, \citet{wang2022fourier} proposed real-time NeRF with a novel  novel Fourier PlenOctree (FPO) algorithm that is designed based on the combination of PlenOctree representation, generalized NeRF, Fourier transformation,  volumetric fusion. Using the PlenOctree to make the fast or real-time NeRF is a good research direction \cite{yu2021plenoctrees}. \citet{guo2022neural} proposed a fast Dynamic NeRF method that focusing on the deformable NeRF, which consists of two main components, adopting a deformation grid for storing the features information of 3D scenes and using the modeling method that contains a color grid and a density.  To increase the convergence of Dynamic NeRF, \citet{liu2022devrf} proposed DeVRF, a method that can render the 3D canonical space and 4D deformation field. In this stage, the characteristic controllable of Dynamic NeRF has drawn more attention. Because controlling the NeRF means the rendering result of the NeRF can be dynamic. \citet{zhang2022controllable} proposed a method that can make the high-fidelity synthesis controllable NeRF with a strategy that contains the parameterization of sequence and motion information.

Focusing on updating and innovating novel implementation methods of Dynamic NeRF is still an important research direction in this stage. The meaning of animatable NeRF is the same as dynamic NeRF in most situations. \citet{yang2022banmo} proposed the Banmo, a method that is different from the normal approach to Dynamic NeRF that is based on the multiple views data and depth information and deformable model of 3D representation, but used the deformable shape model, canonical embeddings and volumetric NeRF. The calibration ability makes the impact of the performance of Dynamic NeRF, \citet{jun2022hdr} proposed the Hdr-plenoxels, a method to increase the dynamic rendering range of Dynamic NeRF by optimizing the calibration with 2D LDR images and a customized tone mapping module. To make a robust Dynamic NeRF system, the approach to get the input data is also important, which means the camera that is used to get the input images or videos is also important. \citet{zhang2022portable} researches the influence for Dynamic NeRF that is brought by the designed multiscopic camera that can generate the input data to make the Dynamic NeRF with any viewpoint at any time. When using the video as the input data, the depth prediction of the reconstruction of the video objects is a significant factor. \citet{novotny2022keytr} proposed the Keytr, a method that designs a key points transfer to improve the estimators of the monocular depth output, by fitting a dynamic point cloud data to the video data. In the application of Dynamic NeRF, assisted driving and autonomous driving is a potential field with great value. \citet{chen2022flow} proposed a method that used the multi-views images that are captured from the camera that are mounted in a car or other driving vehicle, by designing a new training loss and establishing a new dataset, which can render the static and dynamic scenes.

Focusing on making the Dynamic NeRF for human body or human action and motion is still another important research direction in this stage. \citet{zhang2022fdnerf} proposed the Fdnerf to make Dynamic NeRF for the human face emotion by using the few-shot learning. \citet{yao2022dfa} proposed the DFA-NeRF that can render the high-fidelity synthesis of the human face actions with Dynamic NeRF. In this stage using the recovering from the silhouettes to make the surface deformation of human component Dynamic NeRF, to address this problem, \citet{wang2022nerfcap} proposed Nerfcap, a novel method of  performance capturing for Dynamic NeRF, which can easily capture the information from the appearance and geometry transform from the video data. \citet{zhao2022humannerf} also proposed the Humannerf, a Dynamic NeRF representation method that can make the   free-view and high-fidelity 3D synthesis for the Dynamic scenes by using the pixel-alignment feature from the multiple views data and a pose data that is embedded in a non-rigid deformation field. These researches are focusing on the Dynamic NeRF whose rendering scene only has one person, to achieve the Dynamic NeRF for multiple persons, \citet{chao2022mp} proposed the MP-NeRF, a novel Dynamic NeRF method that can render multiple persons in a scene, by using a multiple persons template to make the identification and the motion prior. To improve the rendering efficiency of human components, \citet{xie2022high} proposed UVNeRF, a method that uses the UV-Guided NeRF to make the Dynamic NeRF that is high-fidelity, which consists of the mapping of canonical space, the mapping of texture space and the rendering from the UV-guided NeRF. In the research field of using the Dynamic NeRF to represent the human components, or the research field of 3D modeling, the hair is a difficult component to render. To approach this issue, \citet{wang2022hvh} proposed the Hvh, a method that consists of thousands of primitives and can track the hair on strand level and optimize the 3D scene flow. Rendering the whole head component of a human is an important research topic, \citet{wang2022morf} proposed Morf, a method to make the synthesis of a whole human head with multiple consistent images, which has the controllable and variable identity ability. \citet{lin2022emotional} proposed a semantic NeRF, a method that is designed based on a talking head that is audio-driven, which can render the high-quality 3D video of the human head with the emotion.

\subsubsection{Development of Dynamic in 2023}

In 2023, the development of the research of the Dynamic NeRF has gained a great growth compared with the first two years. As shown in Figure \ref{fig0}, from 2021 to 2023, the number of the researches of Dynamic NeRF is almost doubling every year. Not only the pure basic dynamic type,  more types of Dynamic NeRF has been developed. The developments of Fast NeRF and Editable NeRF are assisted by the evolution of Dynamic NeRF.

Focusing on improving the performance or processing speed of Dynamic NeRF is still an important research direction in this stage. To get higher processing speed of Dynamic NeRF, 
\citet{cao2023hexplane} proposed the Hexplane, a method that uses the 6 planes of the features of rendered objects or scenes that needs to be learned to explicitly represent the 3D synthesis. Some basic Dynamic NeRF that is implemented based on temporally changing of time, will cause the inaccurate and blurry rendering result when the input data is a long  video with complex motion. To address this problem, \citet{guo2022neural} proposed Dynibar, a novel method to make the Dynamic NeRF that uses a designed rendering framework that can synthesize the new viewpoint from the nearby views. Using the light field networks to make the Dynamic NeRF can gain a higher performance compared with the coordinate networks. \citet{yu2023dylin} proposed the DyLiN, a method based on the Dynamic Light Field Network (DyLiN) and is able to deal with the non-rigid deformations. Using the combination of  a canonical NeRF and a motion field can make the Dynamic NeRF, but may cause the problem of inefficient per-scene optimization which needs a long hours time. \citet{geng2023learning} proposed a novel method to make the efficient Dynamic NeRF for humans in minutes with a high visual quality, by designing a voxelized and part-based representation method. Using the method of 4D tensor decomposition can also improve the performance of Dynamic NeRF. \citet{shao2023tensor4d} proposed the Tensor4d, a novel method to make the high-fidelity Dynamic NeRF, by hierarchically  decomposing the 4D tensor into 3 time-ware volumes and 9 feature planes that are compact. To increase the slow speed of rendering and decrease the high cost of storage, \citet{peng2023representing} proposed a novel method to make a high efficient Dynamic NeRF, which is based on using a set of a shallow MLP network to represent the radiance field of each image frame of the input video data, and storing the parameters of the shallow MLP network in 2D grids. 3D Gaussian Splatting \cite{kerbl20233d} is a novel representation technology that has better performance than \cite{muller2022instant} which has an excellent performance in NeRF fast rendering. However, 3D Gaussian Splatting has no advantage in dynamic scenes rendering. To address this problem, \citet{wu20234d} proposed the  4D Gaussian Splatting (4D-GS) that uses a novel explicit representation method which contains both 4D neural voxels and the calculation of 3D Gaussians to get dynamic rendering. To improve the training and rendering speed of Dynamic NeRF, \citet{wang2023mixed} proposed a method that uses the combining idea of combining the static and dynamic voxels process with different designs of networks for the 4D scenes rendering. Keeping the balance between the NeRF rendering speed and quality is an important research direction.  \citet{wang2023neus2} proposed the Neus2, a method that makes use of the CUDA acceleration and lightweight second order derivatives calculation to improve the rendering speed of Dynamic NeRF, which also use a strategy of progressive learning to make the optimization of multi-resolution hash encoding algorithm. \citet{chen2023uv} proposed a novel method that can make the real-time human Dynamic NeRF by separating high-frequency human appearance data from the 3D volume data and encoding the separated data into the data of 2D neural texture stacks.

Focusing on updating and innovating novel implementation methods of Dynamic NeRF is still an important research direction in this stage. \citet{yan2023nerf} proposed Nerf-ds, a method to make the Dynamic NeRF for the specular scene or objects, by reformulating the orientation in the observed space and the Dynamic NeRF function with conduction on the surface position. Transformer as a new neural network is used to design the novel Dynamic NeRF model. \citet{zhu2023cat} proposed the CAT-NeRF, a new design method for Dynamic NeRF, which used 2 layers of Transformer structure as the neural network of the Dynamic NeRF with a designed novel Covariance Loss. This kind of method can precisely model the muscle movements and objects motion in each frame. Making the rendering or representation of a large scale scene is also an important research direction of Dynamic NeRF. \citet{turki2023suds} proposed SUDS, a method to make the scalable Dynamic NeRF that uses the factorization of the scene into 3 separated structures whose data is hash tables and the target signals that is unlabeled and consists of sparse LiDAR, RGB images, 2D optical flow and descriptors. \citet{li2023dynibar} thought the input images data is also important for the performance of Dynamic NeRF, and proposed Dynibar, a method that designs a novel Dynamic NeRF based on the processing of the images, which retains the advantages of modeling complex scenes and having a good view-dependent effects. The Dynamic NeRF of the scenes that have multiple objects is more difficult to implement. \citet{driess2023learning} proposed a novel method to make the Dynamic NeRF for the multiple objects scenes, which is object-centric, automatic encoding and compositional framework, by forcing the latent vectors to be encoded to make the 3D information of NeRF. There are various methods to implement or design a Dynamic NeRF, some researchers have begun to study about what is the most important factor of making a Dynamic NeRF. \citet{park2023temporal} proposed that the temporal interpolation is the most important factor of making Dynamic NeRF, and proposed a  method that uses the spatiotemporal information to train a Dynamic NeRF, and analyzed the method of using the neural networks and using the grids to make the representation. To make editable Dyanmic NeRF, that the key points and topologically varying relationship are considered as important factors.  \citet{zheng2023editablenerf} proposed the Editablenerf, a method to train a dynamic and topologically varying NeRF by processing the information of key points. In this stage, making a Dynamic NeRF solely based on the prior model is limited. To address this issue, \citet{hao2023vt} proposed a vertex-texture latent codes NeRF (VT-NeRF) VT-NeRF, a novel method that used to make the representation of human body, by jointing the latent code and making the combination of the 2D texture latent codes and the vertex latent codes. To design a Dynamic NeRF, the idea of combination is important. \citet{kniaz2023double} proposed Double Nerf, a novel method to make the Dynamic NeRF, which is based on the combination of the static NeRF of the background and the Dynamic NeRF of the animatable objects. \citet{byravan2023nerf2real} also proposed the Nerf2real, a novel method that is based on the design idea of combining the Static NeRF and Dynamic, which can render the high-quality wild scenes. In this stage, making an editable NeRF is more difficult than making a simple dynamic NeRF. \citet{petitjean2023modalnerf} proposed the ModalNeRF, focusing on designing an editable NeRF, which is the first method to make the physically-based edition of the motion of NeRF. For some methods to make the Dynamic NeRF, the deformation field will cause the non-prior and the insufficient information of surface of the rendered objects by the excessive optimization that is under-constrained. To handle this problem, \citet{ma2023high} proposed a novel method to use UV-guided NeRF to design the Dynamic NeRF with sparse input data. Based on the research of UV,  \citet{chen2023uv} also proposed a novel method to make an efficient, real-time and editable rendering or reconstruction of humans, and proposed the idea of UV Volumes that make the design of smaller and shallower NeRF become possible. In the latent space, making a good modeling and reconstruction of the real objects is still a challenge of the inversion issue. \citet{yin2023nerfinvertor} proposed the NeRFInvertor, a novel method to make the high-fidelity Dynamic NeRF by using only one single image. The most popular object of the NeRF research is the normal environment and the human body. For animals, some research groups are focusing on the rendering of normal pets, especially cats and dogs. \citet{sinha2023common} proposed the Common Pets in 3D, a 4D rendering and reconstruction method that is learned from their special video dataset and is used for non-rigid 3D reconstruction. The characteristics of editable and dynamic are similar and relative in NeRF, broadly defined, the editable NeRF is one kind of Dynamic NeRF. Controlling the NeRF is one type of editable NeRF. \citet{lazova2023control} proposed the Control-NeRF, a method that can integrate and control the rendering scenes with the editing of the feature volumes. The editable feature can be also used in the design of real-time Dynamic NeRF. Dynamic NeRF is also used in the area of Physical experiment simulation. \citet{le2023differentiable} proposed a method of using the Dynamic NeRF to implement the physics simulation of moving objects with the study and estimation of the dynamical properties. The moving camera is also an important factor of Dynamic NeRF. Usually the camera of NeRF is considered as slowly moving or static. To study the influence from the camera movement state, \citet{klenk2023nerf} proposed a method that can make the recovery of the NeRF object with a fast moving event camera or objects and a conduction of high dynamic range, which means this method will estimate a NeRF rendering of volumetric scene quickly with a moving camera. As the development of Dynamic NeRF, more and more areas of practical applications have focused on using the Dynamic NeRF to develop the applications for multimedia and Virtual Reality (VR). \citet{song2023nerfplayer} proposed the Nerfplayer, a novel and fast method that can generate the Dynamic NeRF with images from single or multi-camera arrays, which define a point in four dimensions (4D) space into three categories: static, new areas and deforming. Handling the reflected color during the warping of Dynamic NeRF is also important but difficult. To address this problem, \citet{yan2023nerf} proposed the Nerf-ds, a novel method that reformulates the basic  operation function of NeRF on the processing of orientation and surface position. Using the  casually taken monocular images to make the real-time Dynamic NeRF is also an important task. \citet{rosinol2023nerf} proposed Nerf-slam, a method that uses the estimation from the accurate pose data and the data from the depth-maps with associated uncertainty to verify that monocular Simultaneous Localization and Mapping (SLAM) is useful for Dynamic NeRF. SLAM is relative to Structure from Motion (SfM) algorithms. Some SfM technologies are also important for the development of NeRF. Based on the advantages of SfM algorithms, \cite{liu2023robust} proposed a novel method that combines the estimation of both the static and dynamic NeRF, with the data from the camera to get a robust Dynamic NeRF. For the multiple object and scene Dynamic NeRF, using the one whole global Dynamic  NeRF or using the integrated multiple Dynamic NeRF are two main methods to represent multiple scenes. Usually the integrated multiple separated Dynamic NeRF will be more difficult. \citet{driess2023learning} proposed a novel method to achieve a multiple separated Dynamic NeRF, which uses the NeRF decoder to make the 3D data encoding by incorporating the structural priors and making long-term predictions.

Focusing on making the Dynamic NeRF for human body or human action and motion is still another important research direction in this stage. One of further researches of the Dynamic NeRF of human components, especially the human face or human head is to make the controllable Dynamic NeRF.  \citet{athar2023flame} proposed the Flame-in-nerf, a method that focuses on making a controllable Dynamic NeRF for human face, by designing a novel method to make the synthesis of the video about human face, and making a explicitly control mechanism of portrait actions or expressions with an expression in the low dimension. The traditional methods to implement the controllable Dynamic NeRF lack the abilities of combining scenes and making the shape manipulation, etc. To design a more robust controllable Dynamic NeRF, \citet{lazova2023control} also proposed the Control-nerf, a novel method to make the controllable Dynamic NeRF, which only use the optimization of the 3D feature volume that is for specific scene to generate the novel scene synthesis and keep the neural network with the same parameters. \citet{li20223d} also proposed a novel method to make the Dynamic NeRF with 2D observed data for the rigid body and fluid  objects. To improve the rendering efficiency of the Dynamic NeRF of pose of the motions or actions, \citet{tan2023distilling} proposed a method that uses the data of blend skinning and the articulated bones of the rendered object to  represent arbitrary deformations, which can gain a high-quality effect in near real-time. To be more precise in making the Dynamic NeRF of a human body, the clothing of the human body should be paid attention to. \citet{wang2023clothed} proposed a method that uses the tracking of the motion of the human and the clothing of the body as the input data and designing a NeRF field that has two layers. As the development of the Dynamic NeRF for the human component, the efficiency of this kind of NeRF has started to draw more attention. \citet{xu2023avatarmav} proposed the AvatarMAV, a novel method that can fast make the rendering and reconstruction of the human head avatar, by using the weighted multiple 4D tensors concatenation to make the motion aware voxels of the neural network. Except for the making the Dynamic NeRF for the human, taking other animals as the rendered objects is also an significant research direction. \citet{sinha2023common} proposed the Common pets in 3d, a novel method that can make the Dynamic NeRF for the pets, which used a dataset Common Pets in 3D (CoP3D) and proposed the Tracker-NeRF that is the rendering method of learning 4D reconstruction for this dataset. In this research field, the Dynamic NeRF of the action or interaction between the human and objects is also an important research direction. \citet{jiang2023instant} proposed the Instant-NVR, a novel method to make the fast  volumetric human-object Dynamic NeRF, by using the images from a monocular RGBD camera as the input data, with a mechanism that is based on multi-thread rendering and tracking. To make the Dynamic NeRF for human with the data from monocular camera, \citet{yu2023monohuman} proposed the MonoHuman, a framework network to make the animatable NeRF that can generate the high-fidelity rendering and consistent views synthesis under any pose of human, which is using the explicitly information from the the keyframe that is off-the-peg and the bi-directional constraints to make the reasoning of  feature correlations to get a coherent rendering results. Using the traditional NeRF method needs numerous iterations to converge, to address this issue, \citet{xu2023avatarmav} proposed the AvatarMAV, which uses the Motion Aware Neural Voxels to make the fast 3D human head avatar reconstruction. AvatarMAV is also the first method that combines the decoupled expression motion and the canonical appearance. Sometime the capture of the cloth of the human body is significant, \citet{wang2023clothed} proposed a method that can capture clothing motion extract cloth semantics of a human object by using the method of tracking human body motion and the human clothing with a double-layer neural radiance fields. Using the blend skinning and articulated bones can easily build a human object. Based on this, \citet{tan2023distilling} proposed a distilling learning method for the real-time NeRF, which is different from the NeRFs with per-scene optimization that are weak in supporting real-time applications or arbitrary object categories. This method can render the articulated objects that are unseen with a high fidelity at interactive frame rates. Rendering the human object is also influenced by the camera poses. To maintain the high-fidelity rendering under arbitrary novel poses of camera, \citet{yu2023monohuman} proposed the MonoHuman, a novel method that combines the generalizable deformation field which is pose-independent and feature of keyframes to lead the training of the the rendering network. The MonoHuman is focusing on the monocular videos of human objects. \cite{tian2023mononerf} also proposed the Mononerf that is also focusing on the Dynamic NeRF with monocular videos. Therefore the monocular video is an important input data type of Dynamic NeRF. Combining the human prior and NeRF will be a good method to make a Dynamic NeRF of a human body. But this method needs more constraints. \citet{hao2023vt} proposed the VT-NeRF, a method that can improve the gaining of detailed data and integrating the data of vertex latent codes and 2D texture latent codes when processing the surface of the human body, to make a more accurate human body  Dynamic  NeRF.

In the development stage of Dynamic NeRF in 2023, we analyzed three main parts of research directions and applications, including: improving the performance or processing speed, updating and innovating novel implementation and making the Dynamic NeRF for human body or human action and motion. Compared with the first two years, the current development trend of Dynamic NeRF is hot.

\subsection{Review Papers Analysis of NeRF}

This review is the first comprehensive global review that is focusing on Dynamic NeRF. For better understanding the research background and the narrative architecture of the review papers in the similar research field and to write a better review of Dynamic NeRF, ue to The Dynamic NeRF is a subfield of the research field of NeRF, we make a comprehensive summary of the main review papers that are focused on NeRF and represent the comparison of them as two tables as shown in Tables \ref{tb2} and \ref{tb3}. As shown in Tables \ref{tb2} and \ref{tb3}, we selected nine review papers that take up the vast majority of the review papers of NeRF to analyze. The item citation numbers in  Tables \ref{tb2} and \ref{tb3} are the same respectively, with the analysis of focusing and characteristics and content structure of each review paper.

\subsubsection{Review Papers of NeRF in These Three Year}

\begin{table}[htbp]   
\begin{center}   
\caption{Relevant Reviews of NeRF.}  
\label{tb2} 
\tabcolsep=0.1cm
\renewcommand\arraystretch{1.5}
\begin{tabular}{cp{4cm}<{\raggedright}p{8cm}<{\raggedright}}  

\toprule

${\textbf{item/time}}$  &  $\textbf{focus}$  &  $\textbf{characteristics}$    \\

\midrule

\cite{rabby2023beyondpixels}/2023   &   evolution of NeRF  &  comprehensive, evaluation, review recent advances in NeRF and category, explored potentiality  \\

\cite{zhu2023deep}/2023  &   deep review and analysis of NeRF & deep review, main characters, new application innovations, illustrate future opportunities   \\

\cite{molaei2023implicit}/2023   &   medical imaging  & INRs used in various medical imaging, analyse the difficult of processing medical imaging data  \\

\cite{gao2022nerf}/2023  &   3D vision, introduction and review  & comprehensive, papers range from 2021-2023, architecture and application-based taxonomies, NeRF introduction, benchmark comparison of NeRF models  \\

\cite{croce2023neural}/2023  &  digital culture heritage, potential applications  &  preliminary 
critical review, analyse the possible applications in culture heritage domain, comparison of NeRFs and photogrammetry, demonstration   \\

\cite{debbagh2023neural}/2023  &  some recent developments   &  NeRF relative principles introduction, some recent research works about NeRF, not abundant   \\

\cite{xie2022neural}/2022  &   vision computing \newline beyond   &  literature review, context, relative math grounding, 250+ referernced papers, analysis the common components, applications with different problems   \\

\cite{gao2022monocular}/2022  &   monocular dynamic NeRF   &  reality check, dynamic view synthesis (DVS), show discrepancy, define effective multi-view factors (EMFs), introduce 2 new metrics, propose new dataset and experimental protocol \\

\cite{zhan2021multimodal}/2021   &   multimodal image, survey and taxonomy  &   open source survet project, different guidance modalities, describe benchmark datasets, evaluation metrics, current challenges, future directions   \\

\bottomrule
\end{tabular}   
\end{center}   
\end{table}

As shown in Table \ref{tb2}, we compared the published time, research focusing and the characteristics of the existing review papers that are relevant to NeRF. Each item of the review paper is labeled with the corresponding reference number that can be clicked to visit. On the analysis level of published number of the reviews, from 2021 to 2023, the number of the review papers that are relevant to NeRF has grown apace.  Different reviews are focusing on different research fields, such as: evaluation \cite{rabby2023beyondpixels}, medical imaging \cite{molaei2023implicit}, recent developments \cite{debbagh2023neural}, etc. Each review paper that is focusing on the different research topic, has different characteristics of research structure and emphasis. Review \cite{rabby2023beyondpixels} is a review that focuses on the evaluation for different existing NeRF, which means the evaluation and comparison must be two main components of it.  Review \cite{zhu2023deep} focuses on the deep review of the different NeRF research works, which means a deep understanding and analysis of the design and principle of NeRF should be included. Review  \cite{molaei2023implicit} focuses on medical imaging, which emphasizes the different applications of NeRF in the field of medical imaging and analyzing the difficulties of using the NeRF to process the medical data. Review \cite{gao2022nerf} focuses on the normal introduction and the comprehensive review of NeRF, which has the reviewed paper ranges from 2021 to 2023, analyze the architectures and applications of NeRF, and make the relative benchmark comparison. Review \cite{croce2023neural} focuses on a special and different research field of the review of NeRF, the digital culture heritage, which makes the critical analysis and comparison for the applications and researches that are relevant to NeRF. Review \cite{debbagh2023neural} also focus on the recent developments of NeRF, which make the introduction for the relative principles. Review \cite{xie2022neural} focuses on the beyond and novel advanced technologies, which makes the comprehensive introduction about the research background of NeRF, and uses abundant reference papers to support the analysis and content of the review. This review also analyzes the basic principle of designing a NeRF, and the applications of NeRF and the difficulties brought from the implementations. Review \cite{gao2022monocular} focus on the Dynamic NeRF research that is based on the monocular camera, which is the first review paper of Dynamic NeRF in a specific research field but not the first comprehensive review of Dynamic NeRF. This review is focused on using the monocular camera to make the Dynamic NeRF, therefore the content of this review is not comprehensive but focusing on some special fields, such as the DVS and EMFs. Review \cite{zhan2021multimodal} is the last review paper we will analyze, which focuses on using the survey and taxonomy methods to analyze the NeRF that is based on the imaging of multiple models. This review focuses on the engineering projects analysis and evaluation, the current challenges and the further potential research directions.

\begin{table}[htbp]   
\begin{center}   
\caption{Relevant Reviews of NeRF: Content.}  
\label{tb3} 
\tabcolsep=0.15cm
\renewcommand\arraystretch{1.5}
\begin{tabular}{cp{1.5cm}<{\centering}p{11cm}<{\raggedright}} 

\toprule

${\textbf{item/time}}$  &  $\textbf{reference}$    &  $\textbf{content}$   \\

\midrule

\cite{rabby2023beyondpixels}/2023   &  122   & Introduction [1-3] $\rightarrow$ Background [3-4] $\rightarrow$  Advance of NeRF [4-16] $\rightarrow$  Disscussion [16-17]  $\rightarrow$ Conclusion [16-17]   \\

\cite{zhu2023deep}/2023  &  116   & Introduction [2-4] $\rightarrow$ Neural Radiance Fields [4-5] $\rightarrow$  Volumetric Rendering [10-11] $\rightarrow$  Novel View Synthesis [16-17]  $\rightarrow$ Factorizable Embedded Space[11-15]  $\rightarrow$ Multi-view Consistent[15-16] $\rightarrow$  Weighted Importance Sampling[16]  $\rightarrow$  Application Innovations of NeRFs[16-20]  $\rightarrow$  Future[20-22]    $\rightarrow$  Conclusion [22]   \\

\cite{molaei2023implicit}/2023   &   73   & Introduction [1-3] $\rightarrow$ Background [3-4] $\rightarrow$  Clinical Importance [4-5] $\rightarrow$  Taxonomy [5-9]  $\rightarrow$ Comparative Overview[9-11]   $\rightarrow$ Future Work and Open Challenges[11]   $\rightarrow$ Conclusion [11]   \\

\cite{gao2022nerf}/2023  &   200   &  Introduction [1]  $\rightarrow$ Background [2-6] $\rightarrow$  Neural Radiance Field(NERF) [6-17] $\rightarrow$  Applications [17-22]  $\rightarrow$ Discussion[22-23]   $\rightarrow$ Conclusion [23]  \\

\cite{croce2023neural}/2023  &  38   &  Introduction [1-2]  $\rightarrow$ View Synthesis and Neural Rendering [2-3] $\rightarrow$  NeRF Applications in Digital Heritage [3-4]   $\rightarrow$  Case Studies and Early Results [4-6]  $\rightarrow$ Discussion[5]   $\rightarrow$ Conclusion[6]  \\

\cite{debbagh2023neural}/2023  &  10   &  Introduction [1]  $\rightarrow$ Neural Radiance Field [1-3] $\rightarrow$  View Synthesis from fewer Images [3-4] $\rightarrow$  Dynamic and Unconstrained Condition [4-5]   $\rightarrow$ Conclusion [5]  \\

\cite{xie2022neural}/2022  &   250+    &  Introduction [1-3]  $\rightarrow$ Prior Learning and Conditioning [3-5] $\rightarrow$  Hybrid Representations [5-6] $\rightarrow$  Forward Maps [6-8]   $\rightarrow$ Network Architecture [8-9]   $\rightarrow$ Manipulating Neural Fields [8-9]     $\rightarrow$ 3D Scene Reconstruction [9-13]  $\rightarrow$ Digital Humans [13-14]   $\rightarrow$ Generative Modeling [14-15]   $\rightarrow$ 2D Image Processing [15-16]   $\rightarrow$ Robotics [16-17]  $\rightarrow$ Lossy Compression [17]   $\rightarrow$ Beyond Visual Computing [17-18]   $\rightarrow$ Discussion [18-19]   $\rightarrow$ Discussion [18-19]   \\

\cite{gao2022monocular}/2022  &   52   &  Introduction[1-2] $\rightarrow$ Related Work[2-3]  $\rightarrow$ Effective multi-view in a Monocular Video[3-4] $\rightarrow$  Towards Better Experimentation Practice [4-7] $\rightarrow$ Reality Check: Re-evaluating the State of the Art [7-10] $\rightarrow$  Discussion and Recommendation for Future Works [10]  \\

\cite{zhan2021multimodal}/2021   &   304  &   Introduction [1-2] $\rightarrow$ Modality Foundations [2-4] $\rightarrow$  Methods [4-12] $\rightarrow$  Experimental Evaluation [12-14]  $\rightarrow$ Open Challenges and Discussion[14-15]   $\rightarrow$ Social Impacts[15-16]  $\rightarrow$ Conclusion [16]   \\
 
\bottomrule
\end{tabular}   
\end{center}   
\end{table}

\subsubsection{Structure Analysis of the Review Papers of NeRF}

As shown in Table \ref{tb3}, we compared the published time, research focusing published time of the review papers, the number of reference papers that are cited in each review paper, and the structure design of the main content of the listed of each review pacer. To describe the proportion and the distribution of each section of structure of each review in more detail, we use the format: $title[s, e]$ to represent each sub-component of the listed review paper. In this equation, the parameter $title$ represents the title name of the consisted sections of the listed review paper. The parameter $s$ represents the start page number of the specific section, correspondingly, the parameter $e$ represents the end page number of the specific section. With this description equation, it is simple to describe the crossing designed structures and distribution of components of each review paper. We use the symbol $\rightarrow$ to describe the narrative flow of each component of each review paper. 
 
Reviewing other review papers that are relative to the research field of our target review or technical paper, will provide a useful assistance for us to better understand the research background of the specific research field and better design the structure of our review paper. Except for the reference of the main popular structure of the relative review papers, the number of the citation papers and the published time are two important items of our summary, and all are useful for us to better understand the development background of NeRF as the time going and the emphasized research fields of NeRF. The number of the citations can also give a good reference for us to make an available number range of citations. 

Analyzing the structure of each review paper, we can draw the conclusion that the main structure of each review paper can be summarized as including four main components: introduction, background, analysis and conclusion. Therefore this review will follow this structure also.

\section{Basic Method of Original NeRF}

Dynamic NeRF is one kind of the evolution of the basic original NeRF. Some significant principles and design methods of original NeRF are useful for the development and research of Dynamic NeRF. Therefore, in this section, we will analyze the main design idea and method of the original NeRF. One of the most important contributions of the original NeRF \cite{mildenhall2021nerf} is using a 5D vector valued function to represent the expression of a continuous 3D scene.

\begin{figure}[t]
\centering
\includegraphics[width=0.99\columnwidth]{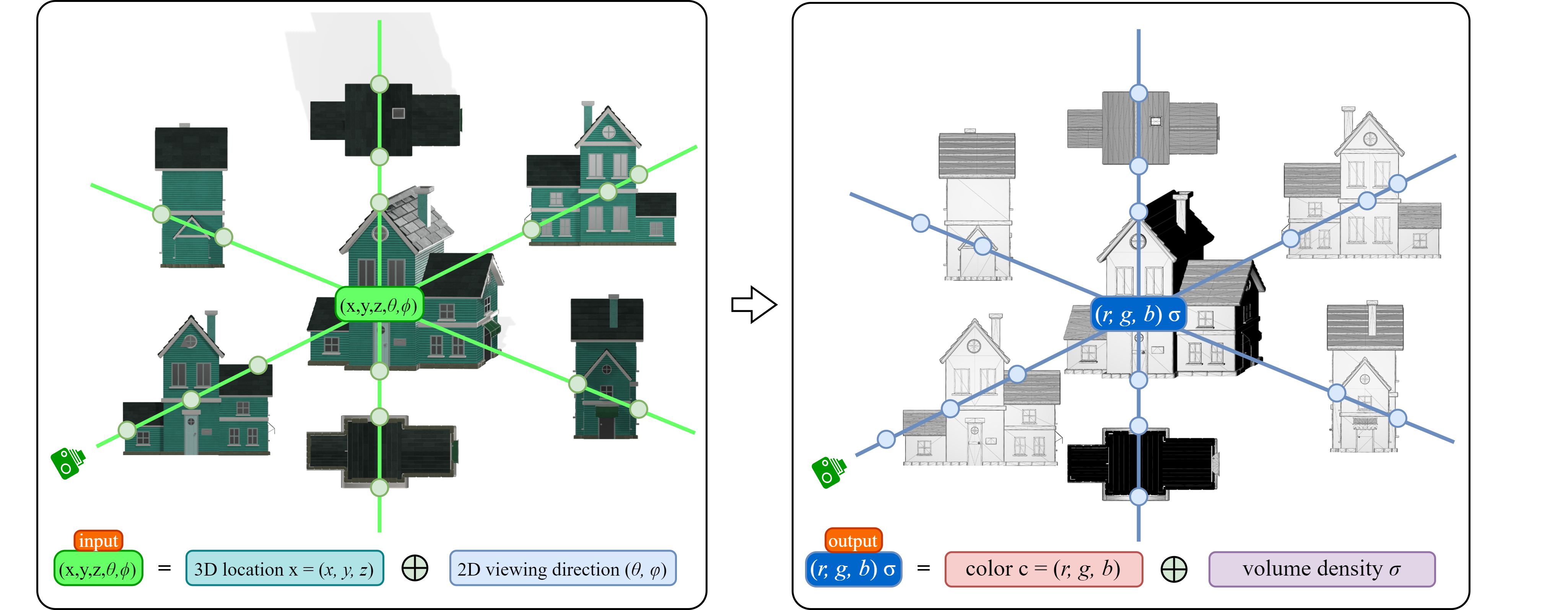}
\caption{Basic Method and Principle of Original NeRF.}
\label{fig1}
\end{figure}

As shown in Figure \ref{fig2}, the input data used for the NeRF is sampled following the camera ray. The camera rays are represented by the green ray in the figure. For each sample point on the ray is consists of two components: the first part is the 3D space location data of the sample point in the scene, which are represented by $x=(x, y, z)$; the second part is the data of 2D viewing direction of the sample point, which are represented by $\theta$ and $\phi$. Therefore, with the combination of the 3D location data and the 2D viewing direction data of the sample points, the input data of the 5D vector valued function can be represented as $(x,y,z,\theta,\phi)$. Subsequently, the input data $(x,y,z,\theta,\phi)$ will be used as input data collection for the 5D vector valued function that is represented as $F_\Theta$. The input of $F_\Theta$ is $(x,y,z,\theta,\phi)$, and the output of $F_\Theta$ is $(r, g, b, \sigma)$.

\begin{equation}
\label{eq1}
\begin{split}
input &= [(\textbf{x},\theta,\phi) = (x,y,z,\theta,\phi)] \xrightarrow{} F_\Theta \\
&\xrightarrow{} [(\textbf{c}, \sigma) = (r, g, b, \sigma)] = output \\
\end{split}
\end{equation}

As shown in Equation\ref{eq1}, for simple expression and description, the 3D location data is represented by $\textbf{x}$, and the 2D viewing direction data is represented by $\textbf{c}$. The function $F_\Theta$ represents the processing of MLP(Multilayer Perceptron) \cite{pal1992multilayer, gardner1998artificial}, which is the main deep learning neural network  framework for NeRF. The result is the output of the MLP function $F_\Theta$ that is a differentiable function, which can use the differential and derivative calculation to get the minimum approximate value by calculating the residual between rendered values that is synthesised and the real world image value of the object. Function $F_\Theta$ represents the 5D vector valued function.

As shown in Figure \ref{fig2}, the output of the MLP function $F_\Theta$ is represented as  $(r, g, b)$. The output consists of two components: one is $\textbf{c}$ which is the colour emitted by the point on the view ray and represented as $(r, g, b)$; the other is $\sigma$ that represents the volume density of the view ray point. Therefore, the output of the 5D vector valued function is combined with the colour information and the volume density information of the point on the camera view ray. The important innovation of NeRF is using the differentiable function $F_\Theta$ to make an approximation and correction with deep learning to approximate the real rendering result by controlling the residual of image data in the real world.

\section{Main Methods Summary}

In this section, the basic and popular method to design and implement the Dynamic NeRF will be stated and discussed, further analysis will also be expanded based on this.

\subsection{Comprehensive Summary}

As shown in Table \ref{tb4}, we make a comprehensive summary for the main method of Dynamic NeRF from 2021 to 2023. The data collection is completed within 10:00 a.m. on January 6th, 2024. To better design Table \ref{tb4}, we make the following definition: $item$ represents the exact cited and analyzed paper, which can be redirected to by clicking on item numbers; $cited$ represents the number of cited times of each paper; $abbrev$ means the abbreviation. In the column of $abbrev$, we use the asterisk $*$ to represent the abbreviated and omitted words in the corresponding positions, to maintain the length of the column; $obj.$ represent the rendering object types, in which the $Nor$ represents the normal types, that means the Dynamic NeRF is designed for general objects rendering, $HB$ means Human Body, $HH$ means Human Head, $HF$ means Human Face, $US$ means Urban Scenes, $CS$ means City Scale and $Obj$ means Objects.

Following, we use two columns to describe the method detail for each paper: the $methods$ column represents the main characteristics of the core method of each paper, the $features$ column represents the main design features of the core method of each paper. The $pul.$ represents the name of the publication journal or conference of each paper and the following abbreviations are made: the CVPR means the IEEE/CVF Conference on Computer Vision and Pattern Recognition, the ICCV means the International Conference on Computer Vision, the WACV means the IEEE/CVF Winter Conference on Applications of Computer Vision, the IRaAL means the IEEE Robotics and Automation Letters, the ITVCG means the IEEE Transactions on Visualization and Computer Graphics and the RobotL means the Robot Learning, the SIGGA means SIGGRAPH Asia. The column $perf.$ represents the performance of the main research result of each paper, with the following abbreviations: the $SOTA$ means the research result is excellent as State-of-the-Art, the $demo$ means the research result has validated the expected functions with demonstrations, the $better$ means the research result has achieved a better result than other specific compared items, the $stable+$ means the research result can achieve more stable result compared with others, the $novel$ means the research method is novel and the result has been validated, the $effect$ means the research result has demonstrated the expected effectiveness of the method, the $100x$ mean the research result has gained a better or faster result that has more than 100 times outperformance than other compared items.  What we can draw from this is, the research result is able to be accepted with a performance that does not achieve the SOTA in the global comparison area but gets a better effectiveness in a specific area.

We use the descending order of the cited number of each paper to sort them and divide them into three year units.

\begin{table}[htbp]   
\begin{center}   
\caption{Main Dynamic NeRF Methods Summary from 2021 to 2023.}  
\label{tb4} 
\tabcolsep=0.15cm
\renewcommand\arraystretch{1.5}
\begin{tabular}{p{0.5cm}<{\centering}p{0.4cm}<{\centering}p{1.2cm}<{\centering}p{0.4cm}<{\centering}p{4.5cm}<{\centering}p{4.2cm}<{\centering}p{0.8cm}<{\centering}p{0.6cm}<{\centering}} 

\toprule

$\textbf{item}$  &  $\textbf{cited}$  &  $\textbf{abbrev}$  &  $\textbf{obj.}$  &  $\textbf{methods}$  &  $\textbf{features}$   &  $\textbf{publ.}$  &  $\textbf{perf.}$  \\

\toprule

2021  \\

\midrule

\cite{park2021nerfies}  &  859  &  Nerfies   &  Nor  &   elastic regularization, deformation
 &   5D, mobile phones, deformable   &  CVPR  &  SOTA   \\

\cite{pumarola2021d}  &  770  &  D-NeRF  &  Nor  &  time as input, two stages process  &  canonical space,  particular time  &  CVPR  &  demo  \\
\cite{li2021neural}  &  482  &  NScene*  &  Nor  &  time-variant continuous function  &  monocular,  camera poses  &  CVPR  &  better  \\
\cite{tretschk2021non}  &  325  &  NRigid*  &  Nor  &  ray bending, rigidity network  &  non-rigid, monocular  &  ICCV  &  stable+  \\
\cite{xian2021space}  &  296  &  Space*  &  Nor  &  constraint time-varying geometry  &  single video, spatiotemporal  &  CVPR  &  better  \\
\cite{peng2021animatable}  &  269  &  Animata*  &  HB  &  canonical NeRF, deformation fields  &  blend weight, skeleton-driven  &  ICCV  &  better  \\
\cite{gao2021dynamic}  &  199  &  Dynamic*  &  Nor  &  time-invariant,time-varying,blend  &  unsupervised, monocular  &  ICCV  &  demo   \\
\cite{ost2021neural}  &  163  &  Neural*  &  Nor  &  encodes transformations+radiance  &  first, decompose, individual  &  CVPR  &  novel  \\
\cite{wang2021learning}  &  77  &  Learning*  &  HH &  3D grid+continuous learned scene  &  combination, maps, global code  &  CVPR  &  SOTA  \\
\cite{yuan2021star}  &  65  &  STaR	 &  Nor  &  jointly optimizing 2 NeRF a frame  &  self supervised, rigid motion  &  CVPR  &  novel  \\

\toprule

2022  \\

\midrule

\cite{li2022neural}  &  183  &  Neural3D*  &  Nor  &  time-conditioned, hierarchical  &  model-free, dynamic setting  &  CVPR  &  SOTA  \\
\cite{hong2022headnerf}  &  145  &  HeadNeRF  &  HH  &  3D proxy, integrating, novel loss  &  parametric, directly control  &  CVPR  &  effect  \\
\cite{yuan2022nerf}  &  123  &  N*-Edit  &  Nor  &  explicit mesh+implicit represent  &  implicit, control, no-re-training  &  CVPR  &  demo  \\
\cite{fang2022fast}  &  98  &  Fast*  &  Nor  &  time-aware voxel, interpolation  &  coordinate deformation network  &  SIGGA  &  SOTA  \\
\cite{wang2022fourier}  &  95  &  Fourier*  &  Nor  &  Fourier PlenOctree(FPO) technique  &  fusion,spatial blending,combine  &  CVPR  &  SOTA  \\
\cite{zhao2022humannerf}  &  94  &  HumanN*  &  HB  &  in-hour scene-specific fine-tuning  &  generalization, no-rigid deform  &  CVPR  &  effect  \\
\cite{huang2022hdr}  &  83  &  HDR-N.*  &  Nor  &  model simplified physical imaging  &  LDR to HDR, ray, mapper, sensor  &  CVPR  &  effect  \\
\cite{zhang2022fdnerf}  &  18  &  FDNeRF  &  HF  &  conditional feature warping  &  few-shot input, arbitrary editing  &  SIGGA  &  SOTA  \\

\toprule

2023  \\

\midrule

\cite{mildenhall2022nerf}  &  207  &  N*Dark*  &  Nor  &  rendering raw output from results  &  dark, linear images, manipulate  &  CVPR  &  better  \\
\cite{li20223d}  &  88  &  3DNeural*  &  Nor  &  combines NeRF and time learning  &  autoencode, viewpoint-invariant  &  RobotL  &  demo  \\
\cite{cao2023hexplane}  &  87  &  HexPlane  &  Nor  &  six planes, fusing vectors, tiny MLP  &  regress colors, train via volume  &  CVPR  &  100×  \\
\cite{song2023nerfplayer}  &  66  &  Ne*Player  &  Nor  &  decompose 4D space, probabilities  &  static, deforming, and new areas  &  ITVCG  &  SOTA  \\
\cite{driess2023learning}  &  46  &  Learn*Mu*  &  Nor  &  object-centric auto-encoder, priors  &  compositional, latent vectors  &  RobotL  &  stable+  \\
\cite{li2023dynibar}  &  39  &  DynIBaR  &  Nor  &  image-based, encoding with MLPs  &  long videos, unconstrained  &  CVPR  &  SOTA  \\
\cite{lazova2023control}  &  38  &  Control*  &  Nor  &  edit independent feature volumes  &  controllable, editable, combine  &  WACV  &  demo  \\
\cite{wang2023neus2}  &  37  &  NeuS2  &  Nor  &  lightweight multi-resolution hash  &  CUDA,surfaces,fast,incremental  &  ICCV  &  SOTA  \\
\cite{shao2023tensor4d}  &  34  &  Tensor4D  &  HB  &  efficient 4D tensor decomposition  &  decompose, projecting, 4D fields  &  CVPR  &  effect   \\
\cite{turki2023suds}  &  20  &  SUDS  &  US  &  separate hash table, target signals  &  large-scale urban, 2D descriptor  &  CVPR  &  SOTA  \\
\cite{wang2023mixed}  &  20  &  Mixed N*  &  Nor  &  static+dynamic voxels, different net  &  fast, lightweight static, variation  &  ICCV  &  better  \\
\cite{geng2023learning}  &  15  &  Learning*  &  HB  &  part-based voxelized,different parts  &  2D motion parameterization  &  CVPR  &  100×  \\
\cite{zhu2023latitude}  &  15  &  LATITUDE  &  CS  &  truncated Dynamic low-pass filter  &  regressor,residual,coarse-to-fine  &  ICRA  &  effect   \\
\cite{le2023differentiable}  &  14  &  Different*  &  Obj  &  differentiable simulating pipeline  &  continuous density field, friction  &  IRaAL  &  demo   \\
\bottomrule
\end{tabular}   
\end{center}   
\end{table}

\subsection{Comparison and Analysis}

To better understand the development of Main Dynamic NeRFs from 2021 to 2023, we make a further detailed analysis for summary of Table \ref{tb4}. 

\begin{figure*}[htbp]
\centering
\subfloat[global]{\label{fig2a} \includegraphics[width=0.32\linewidth]{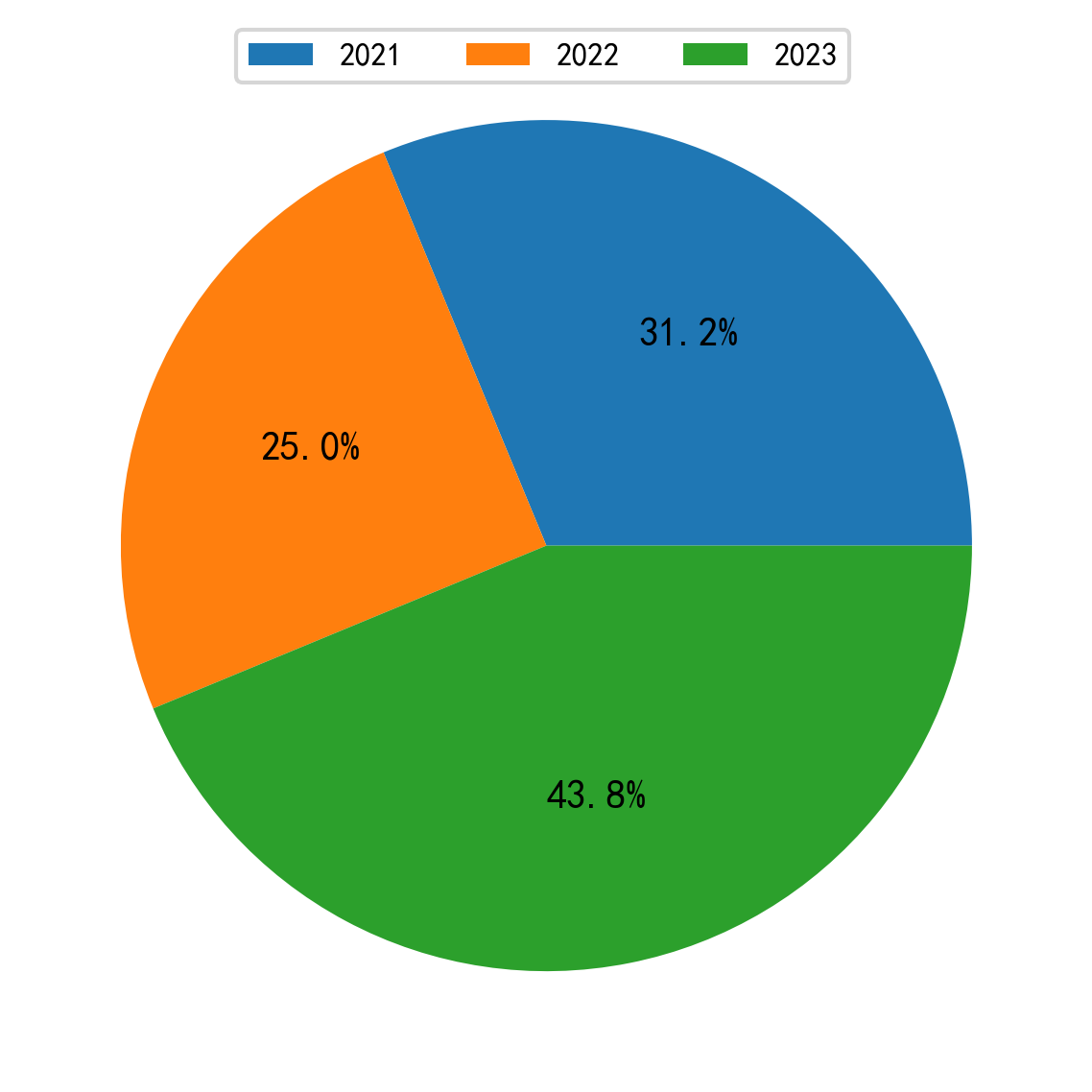} }  \hspace{0pt}
\subfloat[obj.]{\label{fig2b} \includegraphics[width=0.32\linewidth]{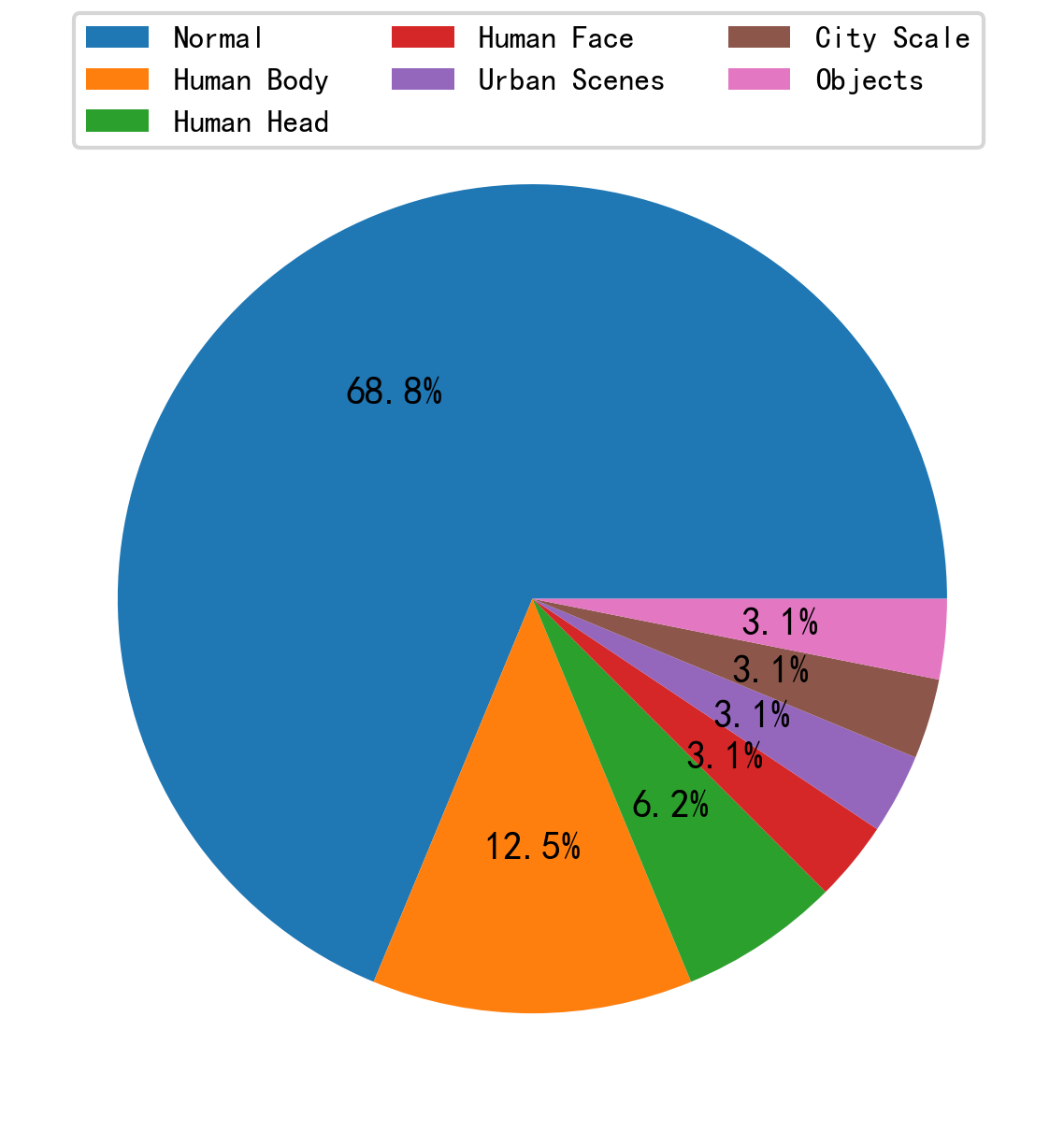}}  \hspace{0pt}
\subfloat[pul.]{\label{fig2c} \includegraphics[width=0.32\linewidth]{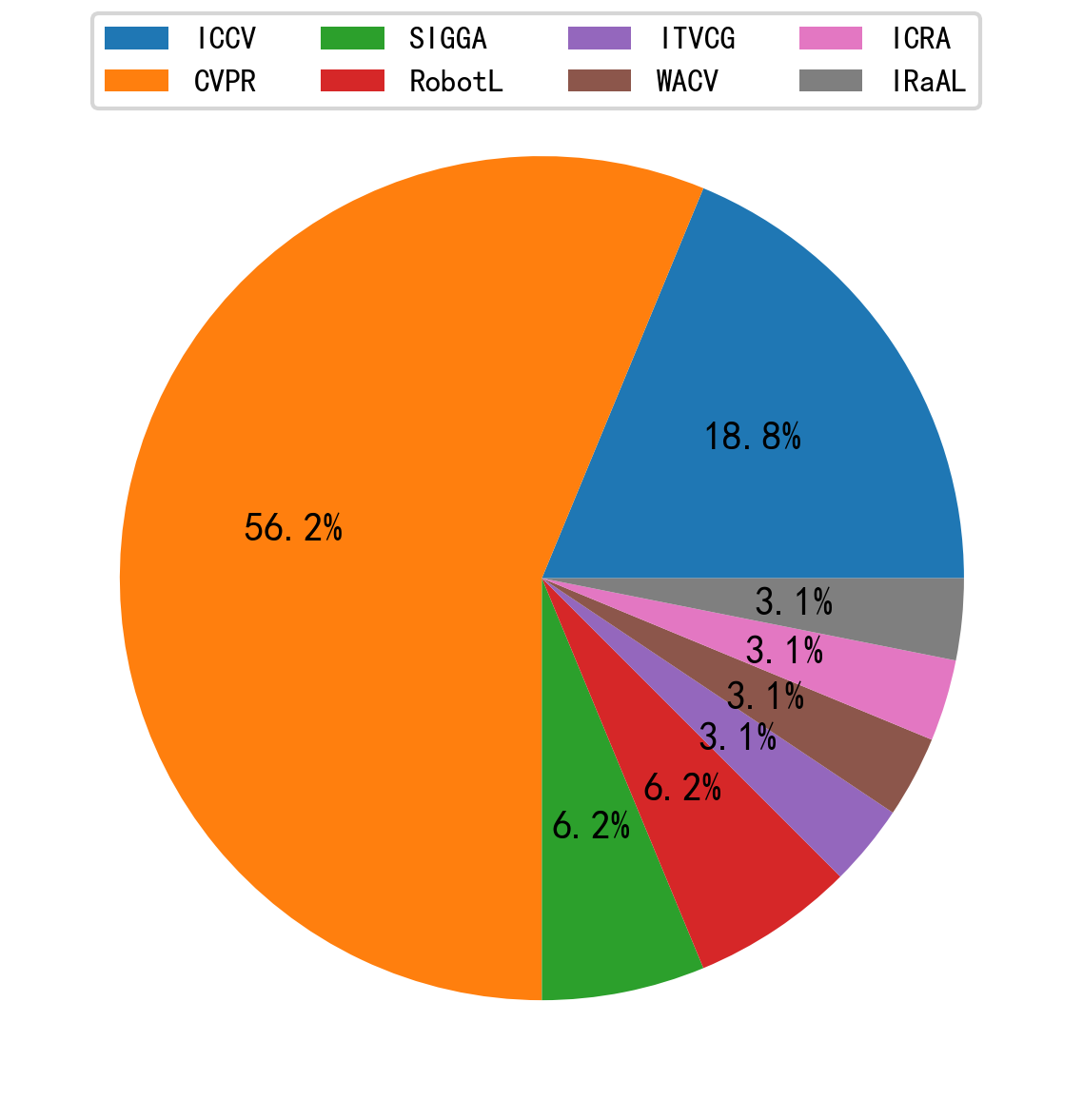}}  
\\
\subfloat[papers]{\label{fig2d} \includegraphics[width=0.3\linewidth]{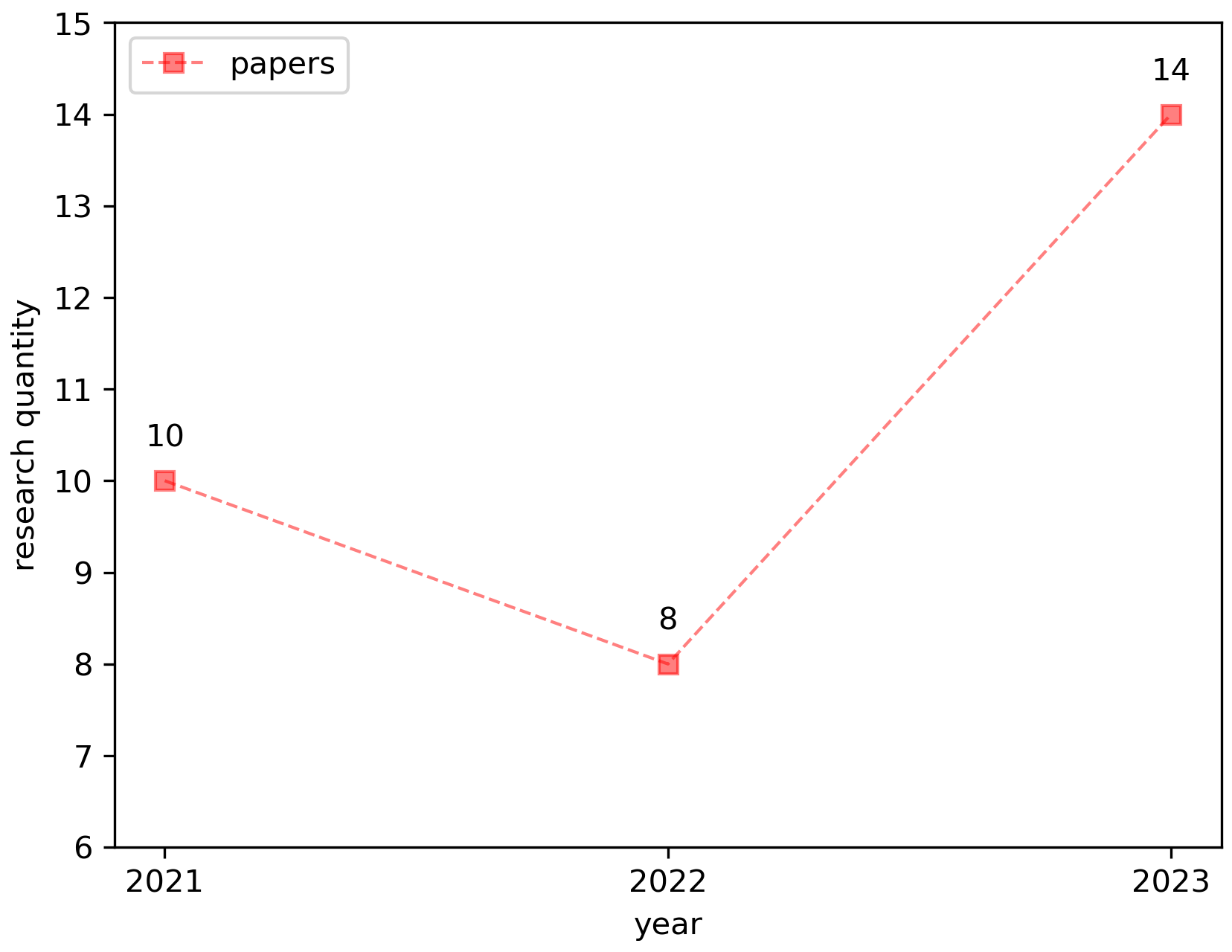}}  \hspace{0pt}
\subfloat[citations]{\label{fig2e} \includegraphics[width=0.3\linewidth]{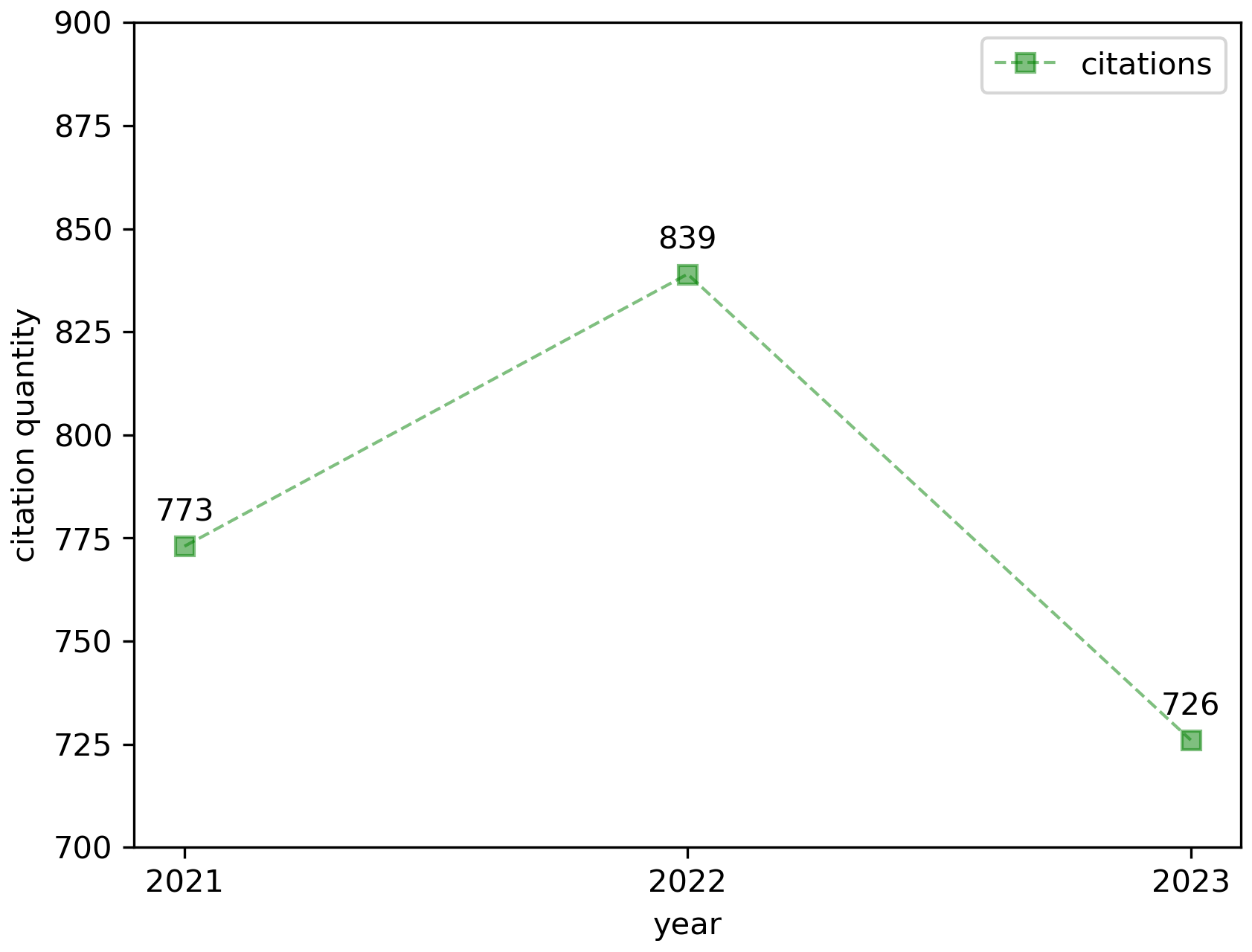}}  \hspace{0pt}
\subfloat[citations per paper]{\label{fig2f} \includegraphics[width=0.3\linewidth]{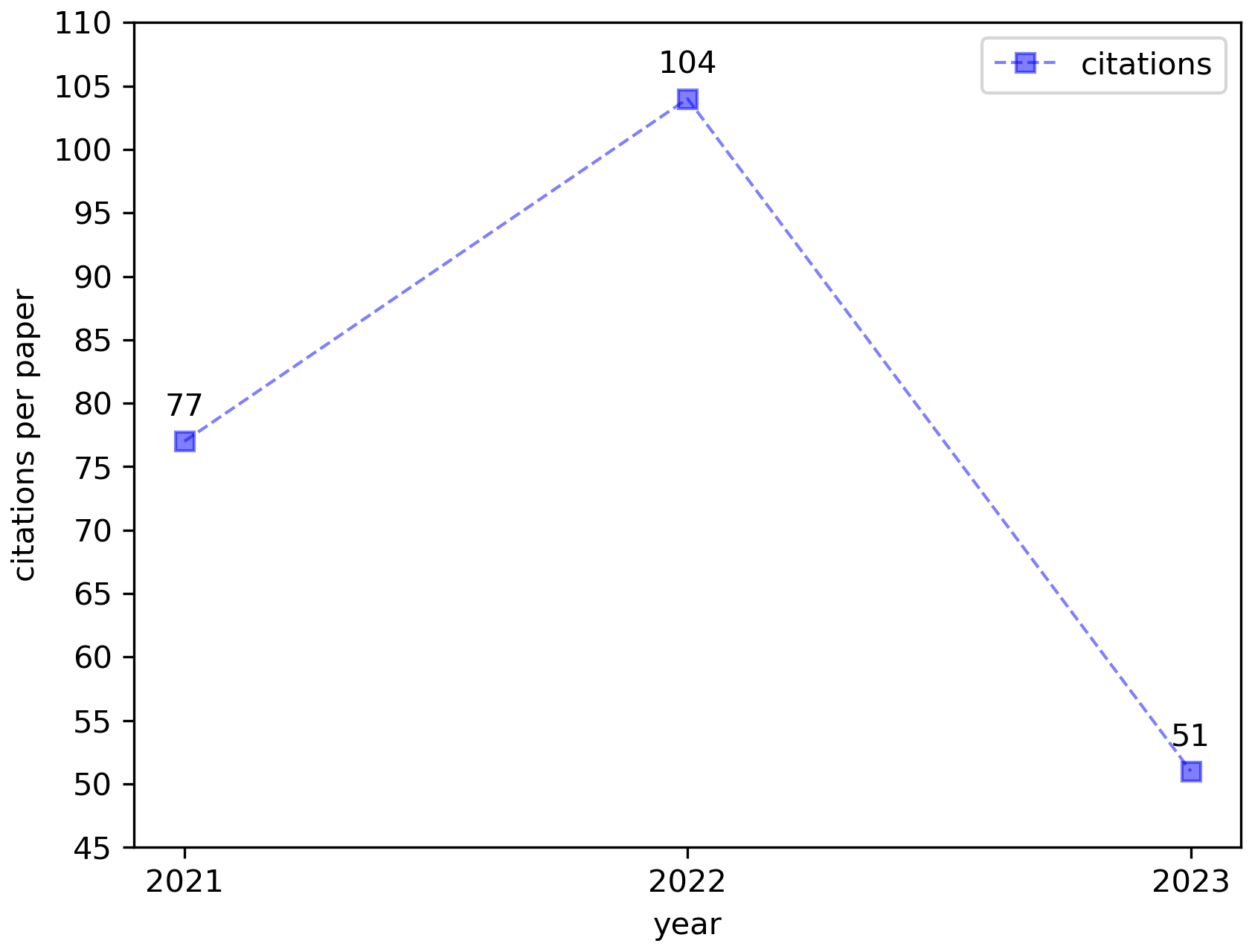}}
\caption{Comparison and Analysis of Dynamic NeRF from 2021 to 2023.}
\end{figure*}

First, we make the classification and proportion analysis of Dynamic NeRF from 2021 to 2023. As shown in Figure \ref{fig2a},  the proportion of analyzed papers in the Table \ref{tb4} has been greatly expanded in 2023, although the number in 2022 is less than 2021. For the rendering object types, as shown in Figure \ref{fig2b}, the most popular object type is the normal type, which means the Dynamic NeRF for general objects is the mainstream. What should be noted is that the Dynamic NeRF research that focuses on the human individual or human components is the second biggest research direction. Big large scale Dynamic NeRF is also a popular research direction. For the publication journals or conferences analyzes, as shown in Figure \ref{fig2c}, the impressive fact is that more than half of the analyzed papers are published in CVPR, the second proportion is held by ICCV. Both CVPR and ICCV are the conferences of most popular and primary conferences that are related to research of computer vision. In other words, an available suggestion is that the CVPR and ICCV are two recommended conferences for the publication of good research of Dynamic NeRF.

Subsequently, we make the development and trend analysis of Dynamic NeRF from 2021 to 2023. As shown in Figure \ref{fig2d}, although the analyzed paper's number of Dynamic NeRF in 2022 has reduced compared to 2021, the number has grown in 2023. Although the number of papers in 2022 is the least, the number of total citations is the largest, as shown in Figure \ref{fig2e}. Due to the papers that are published in 2023 are the latest results, the citation number is the least but the gap is not large compared with others. As shown in  Figure \ref{fig2f}, the research results that are published in 2022 have gained more individual citation rate, which means those papers have higher quality to draw more research attention.

\section{Significant Methods Analysis}

In this section, we will discuss the different approaches to achieve Dynamic NeRF. The method classification analysis is used in the statement of this section.

\subsection{Time as an Additional Input}

\begin{figure}[t]
    \centering
    \includegraphics[width=0.99\columnwidth]{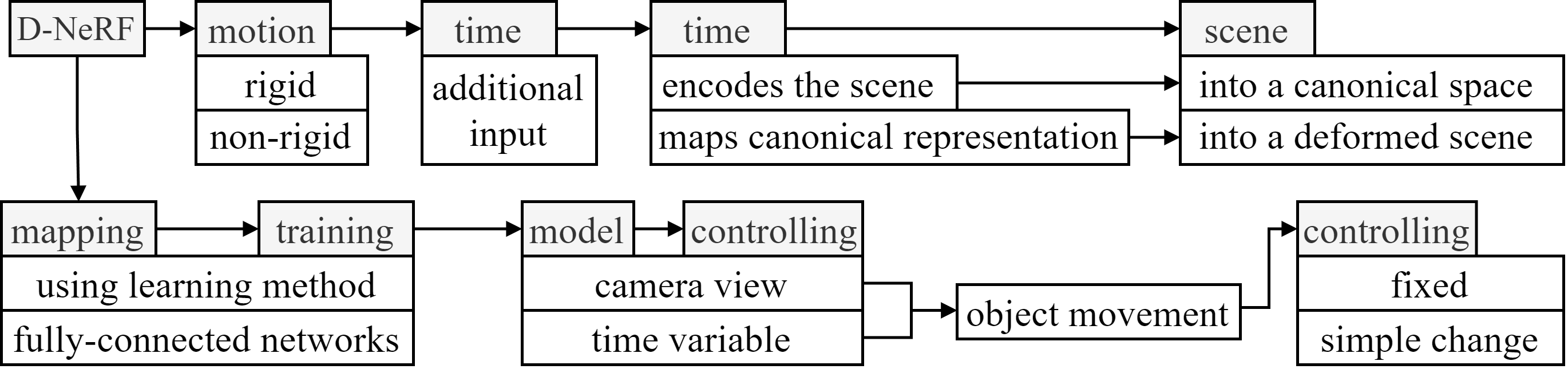}
    \caption{Design main structure of D-NeRF on each main function of the structure.}
    \label{fig2}
\end{figure}

The most classic method to implement the Dynamic NeRF is adding the time as a additional input or domain of the original static NeRF system (e.g.: D-nerf\citet{pumarola2021d}). As shown in Figure\ref{fig2}, the D-NeRF processed objects can be rigid and non-rigid types. Cored idea is adding time as an additional input in machine learning network. Adding the time serves two purposes: encoding the scene into a canonical space; and mapping the canonical representation into a deformed scene. Mapping the scene in training has two features: using the learning method and using the fully-connected networks. After achieving the training of mapping with that addition of time data, it is available to control the model in two main two main aspects: controlling the camera view angle and controlling the time with corresponding scene representation. Therefore, this kind of controlling has two main short-backs: the controlling transformation is fixed, and the changing is simple. In general, this is one of the first Dynamic NeRFs. 

\begin{figure}[t]
    \centering
    \includegraphics[width=0.99\columnwidth]{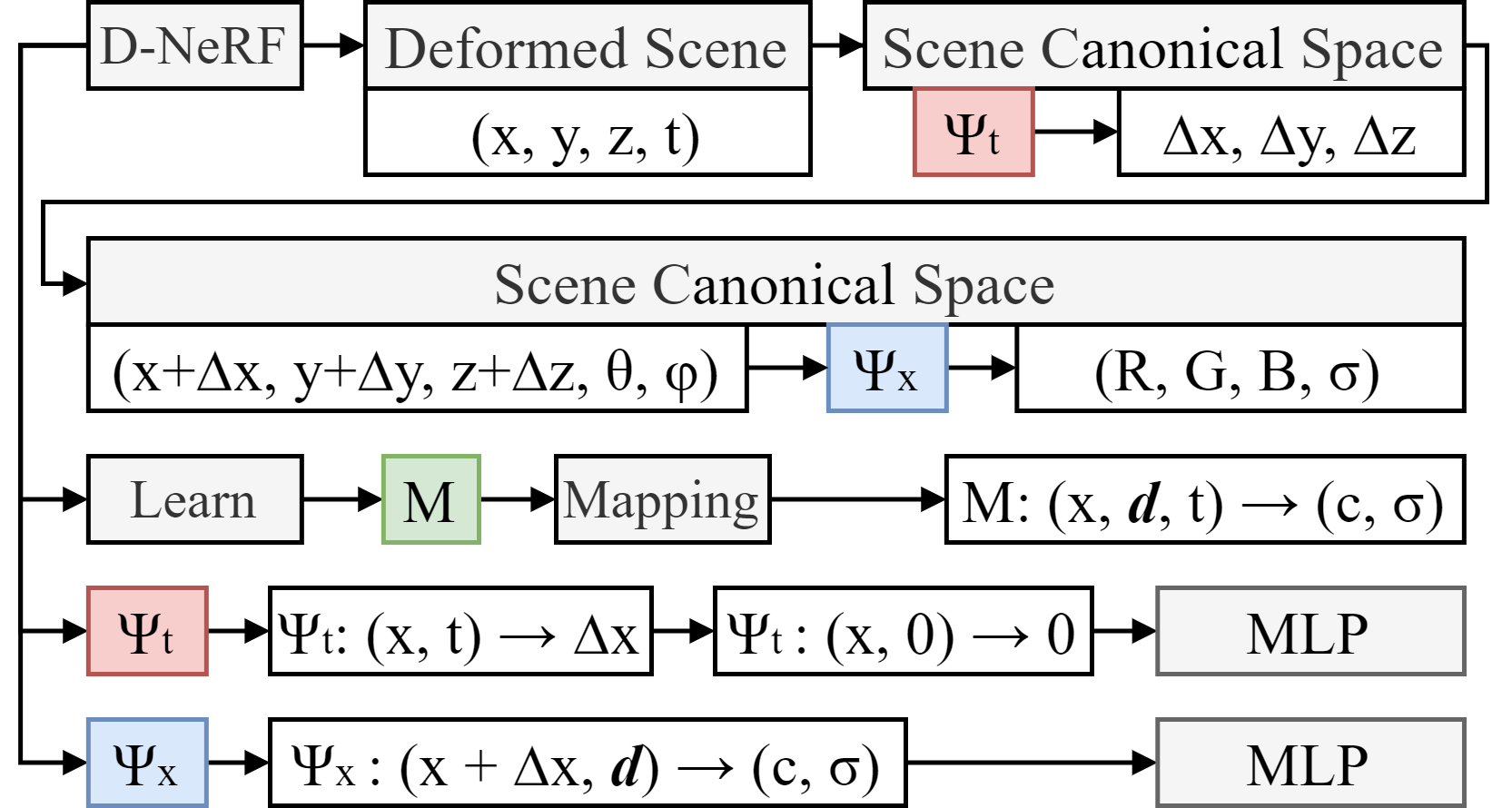}
    \caption{Design and functions of $t$ canonical deformation MLP $\Phi_{t}$ and encoding MLP canonical configuration deformation $\Phi_{x}$.}
    \label{fig3}
\end{figure}

As shown in Figure\ref{fig3}, two process parts consist the main operation of D-NeRF Model. $\Phi_{t}$ is a deformation network and $\Phi_{t}$ is a canonical network. Using the $x = (x, y, z)$ to represent a 3D point in the scene. Color of point $x$ is represented as $c = (r, g, b)$. Volume density of the point on a time instant $t$ is $\sigma$, when the view direction is $\mathbf{d}=(\theta,\phi)$. The input variables are $(x, y, z, t)$ in deformed scene. Mapping $M$ can be described as $M:(x, \textbf{d}, t) \to (c, \sigma)$. Then the input data is processed by canonical space $\Psi_t$. $(\Delta x, \Delta y, \Delta y)$ represent the coordinates deformation value in direction $x, y, z$ respectively. Making the differential computation when $\Psi_t:(x, t) \to \Delta x$, i.e.: $\Psi_t:(x, 0) \to 0$. Calculating the a common canonical space anchor of $\Psi_t$ with a MLP network. Following, assigning the emitted color $c$ and volume density $\sigma$ with a canonical configuration deformation $\Psi_x:(x+\Delta x, \textbf{d}) \to (c, \sigma)$, using the MLP network.

\begin{figure}[t]
    \centering
    \includegraphics[width=0.99\columnwidth]{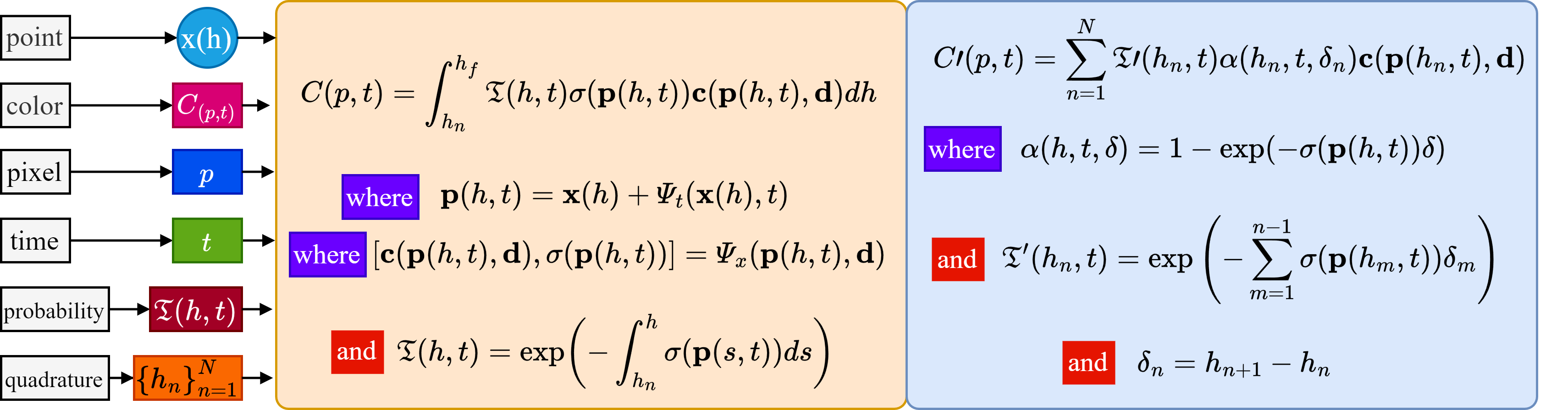}
    \caption{Calculating of volume rendering color $C$ of D-Nerf of pixel $p$ at time $t$ with point $x(h)$.}
    \label{fig5}
\end{figure}

After gaining the calculation of $\Phi_{t}$ and $\Phi_{x}$, the next step is calculating detailed volume rendering. First defining the rendered point as $x(h) = o+h\textbf{d}$. Center of the projection ray is represented as $o$, which terminal pixel is $p$. Bounds $h_n$ and $h_f$ are the near and far bounds respectively. Finally, it is able to calculate the expected rendering color $C$ of pixel $p$ at time $t$ by the calculation as shown in Figure\ref{fig5}. In this calculation, it is need to calculating the accumulated probability of the projection ray $\mathfrak{T}'(h,t)$ that is emitted from bound $h_n$ and bound $h_f$ when it missed hitting any other particle.

As shown in Figure\ref{fig5}, using the pixel $p$ and the corresponding time $t$ can define a rendering color $C(p, t)$ that is a definite integral from $h_n$ to $h_f$. The calculating needs to satisfy the condition equations between $p$ and $\Psi_{t}$ or $\Psi_{x}$, meanwhile, the projection ray $\mathfrak{T}(h,t)$ should be calculated. The density $\sigma$ and color $C$ is gained by the prediction calculation of MLP network $\Psi_{x}$. The equations of estimating the colors $C'(p,t)$ of the chosen pixels is shown in the right blue box. Adding $\delta$ in the the $p(h, t)$ to calculate a satisfaction condition. Analogously, calculating the $\mathfrak{T}'(h,t)$ with $\delta_n$. After the calculating of rendering color of each point pixel, the following step is training the learning network of the whole model. The D-NeRF model is trained with a single GTX-1080(Nvidia) for 2 days.

\subsection{Deforming Ray Bending}

The second main method we want to introduce and analyze is the implementing the scene deformation by using the ray bending. Non-rigid neural radiance fields is a representative research project that makes use of the feature of ray bending \cite{tretschk2021non}. Using this kind of method can achieve some special complex rendering effects, for example, the delay showing effect likes bullet-time (i.e.: the time slice effect or time freeze effect calling in the previous time). In this method, the main outstanding advantage is implementing the ray bending and rigidity network learning need not explicit supervision. The research work \cite{tretschk2021non} is abbreviated as NR-NeRF by the authors. As similar with many other NeRF research works, NR-NeRF still has a general weak short-back that will take a whole NeRF rendering network for a unit or independent object or scene rendering. In the future, comprehensive and combined big NeRF rendering from the multiple split independent sub-object renderings will be a main research trend, which is more difficult and complex.

\begin{figure}[t]
    \centering
    \includegraphics[width=0.99\columnwidth]{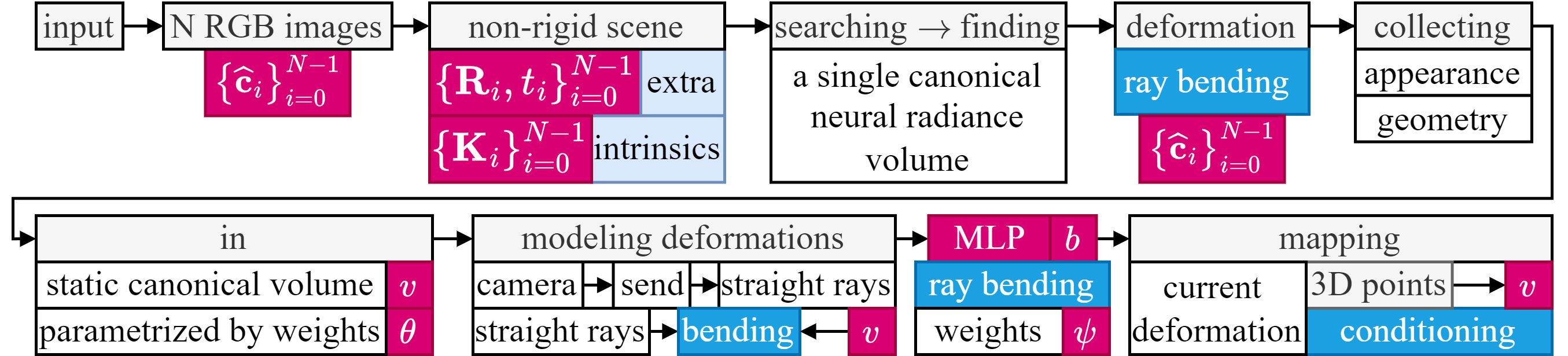}
    \caption{Main processing and rendering operations of NR-NeRF from input image to mapping.}
    \label{fig6}
\end{figure}

As shown in Figure\ref{fig6}, the count of the input image of NR-NeRF is not a single but multiple images, which is represented by $\{\hat{\mathbf{c}}_i\}_{i=0}^{N-1}$. The input images are from the non-rigid scene that has two kinds of inherent features: the extrinsics ${\mathbf{R}_i, t_i}_{i=0}^{N-1}$ and the intrinsics ${\mathbf{K}_i}_{i=0}^{N-1}$. Subsequently, a single canonical neural radiance volume, that is deformed via ray bending, will be searched and found, which is used to render the color in $\{\hat{\mathbf{c}}_i\}_{i=0}^{N-1}$. In the main calculation and processing of rendering the colors, the appearance and geometry information of the static canonical volume that is represented by $v$ will be weighted and parameterized with weights $\theta$. Following, entering the step of modeling the deformation, which is an important strategy that changing the traditional processing method to novel machine learning method. First, collecting the information from the straight rays that are sent from the camera, followed by defining a deformation $v$ that presents the ray bending deformation processing of the straight rays. Following, adding the MLP design $b$ on the ray bending with the weight $\psi$ as input parameters. Finally, handling the mapping processing that is implemented on static canonical volume $v$. In the mapping processing, a main operation is gaining the form from a designed auto-decoded latent code ${\mathbf{I}_i}_{i=0}^{N-1}$. To here, the whole processing is completed.

\subsection{Skeleton-driven Deformation}

3D reconstruction and representation of human body is sustaining hot research topic, due to the wide range of practical application value. Not only for the 3D modeling, 3D video but the 3D game design, human-like object modeling is also important, as well as in the dynamic NeRF area. As a representative research project, \citet{peng2021animatable} proposed a novel method developed based on traditional human armature animation deformation, which is abbreviated as Skeleton-NeRF in this paper. Skeleton-NeRF is also trained based on blending weight field that produce deformation field.   

\begin{figure}[t]
    \centering
    \includegraphics[width=0.99\columnwidth]{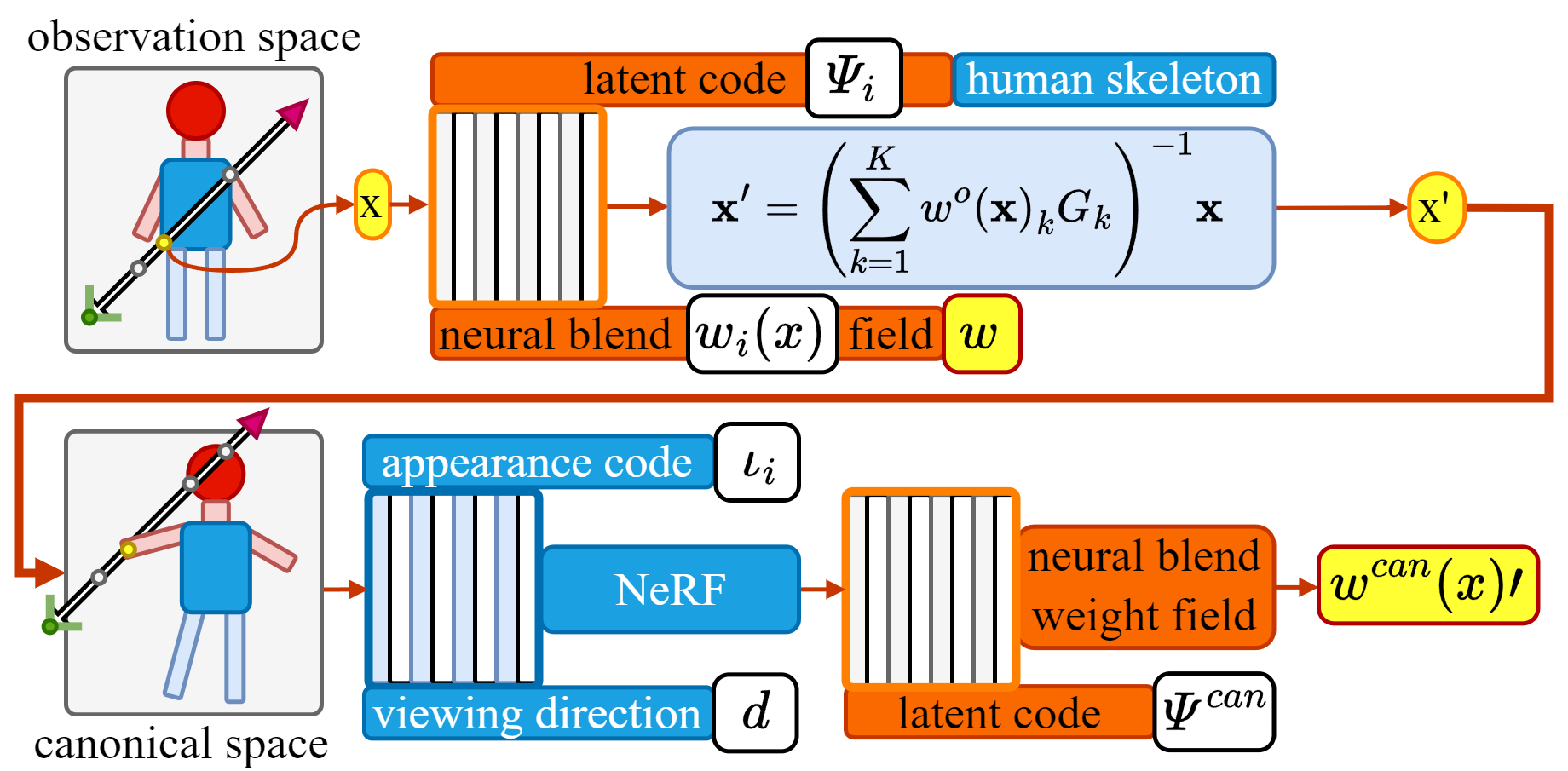}
    \caption{Main processing and rendering operations of Skeleton-NeRF that is based on skeleton-driven deformation.}
    \label{fig7}
\end{figure}

As shown in Figure\ref{fig7}, in Skeleton-NeRF, when rending the scene, each frame $i$ will be processed by extracting a query point $x$ in the observation space. Each query point $x$ is inferred to gain its blend weight that is represented as $w_i(x)$. Inferring the weight $w_i(x)$ of point $x$ is using a neural blend wight field. The neural blend field is conditioned by a latent code $\Psi_i$. The weight is abbreviated as $w$. Subsequently, with the equation $\mathbf{x}^{\prime}=\left(\sum_{k=1}^Kw^o(\mathbf{x})_kG_k\right)^{-1}\mathbf{x}$, we can calculate the corresponding transformed point $x'$ in the canonical space, from the input point $x$. Parameter $d$ represents the viewing direction of the observation space. Following, enter the processing in the canonical spec. The appearance code that is represented by $\iota_i$ will be one of the input variables of the NeRF network. The other input variable of the NeRF network is the viewing direction $d$ in observation-space. After the processing of NeRF network, with the NeRF prediction, the output are the rendered volume, density and color. Finally, using the neural blend weight field $w^{can}(x')$ network to learn the animation of NeRF, which is implemented at the canonical space, to gain the dynamic rendering.

\section{Conclusion}

This review is a comprehensive technology and statement review. In this review, we discussed and analyzed a large number of research projects and results in the area of Dynamic NeRF, which covers almost all of the important results of Dynamic NeRF from different sub-domains. To better present the statements and comparisons of the Dynamic NeRF in various of dimensions, we proposed some novel presentation method on figure and table. In the sections of analyzing the concrete implementation methods of Dynamic NeRF, we used a novel method to combined the words-equations and diagrams to show the implementation process, which will be more simple understood. The key design of Dynamci NeRF, compared with the static NeRF, is adding the deformation factor or domain into the previous static NeRF network. Howeve, those Dynamic NeRF is not supporting the re-editable control or edition, which should be payed more attention to update the methods in the further research or in the future. Most of the researches of Dynamic NeRF is useful for the research of editable or re-editable NeRF in the future. In some dimensions, some principles of Dynamic NeRF are even useful for the design of Gaussian Splatting research projects, which are our next key study object.


\bibliographystyle{ACM-Reference-Format}
\bibliography{sample-sigconf}










\end{document}